%% file: ijcai26.tex
\documentclass{article}
\usepackage{ijcai26}

\usepackage{times}
\usepackage{soul}
\usepackage{url}
\usepackage[hidelinks]{hyperref}
\usepackage[utf8]{inputenc}
\usepackage[small]{caption}
\usepackage{graphicx}
\usepackage{amsmath}
\usepackage{amsthm}
\usepackage{booktabs}
\usepackage{algorithm}
\usepackage{algorithmic}
\usepackage[switch]{lineno}
\usepackage[algo2e,ruled,noend,linesnumbered]{algorithm2e}
\usepackage{color,soul}
\usepackage{xcolor}
\usepackage{multicol,multirow}
\usepackage{subcaption}
\usepackage{bm}

\newtheorem{defn}{Definition}

\title{A Lightweight Traffic Map for Efficient Anytime LaCAM*}

\author{
Bojie Shen
\and
Yue Zhang
\and
Zhe Chen
\And
Daniel Harabor\\
\affiliations
Monash University, Faculty of Information Technology \\
\emails
\{bojie.shen, yue.zhang, zhe.chen, daniel harabor\}@monash.edu
}

\begin{document}

\maketitle

\begin{abstract}
Multi-Agent Path Finding (MAPF) seeks collision-free paths for teams of agents and has a wide range of practical applications. LaCAM*, an anytime configuration-based solver, currently represents the state-of-the-art. Recent work has explored using guidance paths to steer LaCAM* toward configurations that avoid traffic congestion, thereby improving solution quality. However, existing approaches rely on Frank–Wolfe–style optimisation to repeatedly invoke single-agent search before executing LaCAM*, which creates a large computational overhead in large-scale problems. The guide path is also static, which is only helpful for finding the first solution in LaCAM*.
To overcome this problem, we propose a new approach that exploits LaCAM*’s ability to construct a dynamic, lightweight traffic map during LaCAM*'s search.
Experiments show that our method achieves higher solution quality than state-of-the-art guidance-path approaches in two variants of MAPF problems. 

\end{abstract}

\input{section/introduction}

\input{section/preliminary}
\input{section/algorithm}
\input{section/experiment}

\bibliographystyle{named}
\bibliography{ijcai26}

\end{document}

%% file: section/introduction.tex
\section{Introduction}

Multi-Agent Path Finding (MAPF) is a fundamental coordination problem in robotics 
and artificial intelligence. The problem asks computing collision-free paths for a team 
of agents navigating a shared environment~\cite{stern2019multi}. 
This problem lies at the heart of numerous large-scale industrial applications, most notably in 
automated warehouse~\cite{wurman2008coordinating}, computer games~\cite{silver2005cooperative}, and railway scheduling~\cite{flatland}. 
In these applications, hundreds or even thousands of moving agents must operate simultaneously to optimise a system objective. In such settings, optimal solvers typically struggle to scale as the number of agents grows~\cite{shen2023tracking,bcp-mapf,lazycbs,cluster_reasoning,li2021eecbs}. 
Consequently, real-world deployments often rely on 
suboptimal, unbounded approaches capable of generating feasible plans within tight, real-time computational limits~\cite{okumura2023engineering,okumura2022priority,chen2024traffic,JiangSoCS24}.

To address these scalability challenges, rule-based algorithms like Priority 
Inheritance with Backtracking (PIBT)~\cite{okumura2022priority} and search-based frameworks like LaCAM/LaCAM*~\cite{okumura2023lacam,okumura2023improving} have emerged.
These methods typically rely on ``myopic'' heuristics to guide the search, such as individual shortest distances, which blindly steer agents into high-traffic areas, resulting in severe congestion and poor solution quality. 
To mitigate this, recent works introduce guidance mechanisms that steer agents away from potential bottlenecks. 
For instance, 
Space Utilization Optimization (SUO)~\cite{SUO} precomputes spatially dispersed paths to diversify agent movements, helping to guide LaCAM/LaCAM* to avoid spatial congestion~\cite{okumura2023engineering}. 
Traffic Flow Optimisation~\cite{chen2024traffic} approaches adapt ideas from traffic assignment problems, using iterative replanning to optimise congestion penalised guide paths (Frank-Wolfe-style optimisation) towards a User Equilibrium, where no agent can improve its travel time by switching routes. 
These paths then replace the default distance heuristic to guide PIBT.
More recently, learning-based methods have been 
proposed to further optimise guidance policies or graphs to capture dynamic traffic patterns~\cite{GGO,OnlineGGO}.

Despite their success in improving solution quality, these guidance-based approaches suffer from significant limitations regarding computational efficiency and adaptability. 
Methods like Traffic Flow Optimisation and SUO typically employ repeated computationally expensive pathfinding operations for every agent to reach a traffic equilibrium before the actual search begins. 
This introduces substantial setup overhead, often requiring dozens of seconds of pre-computation for large-scale instances,
which undermines the real-time performance critical for online applications. 
Similarly, learning-based approaches often incur high training costs or require complex neural network updates that can be computationally intensive to execute online.
Furthermore, these learnt policies are often highly specific to the training environment. Applying them to a new map topology or domain typically necessitates an extensive training process involving thousands of simulations. 

To address these limitations, we propose a novel online mechanism called the Lightweight Traffic Map (LTM), designed to integrate seamlessly with the LaCAM framework. 
Instead of relying on an expensive, two-stage offline optimisation phase, our approach exploits LaCAM's inherent ability to rapidly sample configurations during the search itself. 
By recording the historical solution data from the anytime search process, we construct a dynamic weight map to represent the congestion in real-time. 
This traffic map is then used to bias subsequent search iterations and updated in real-time, effectively steering agents away from congested regions without incurring the computational overhead of repeated single-agent path replannings.
We evaluate the proposed approach on a wide range of standard benchmark maps. Experimental results demonstrate that our method converges faster and produces higher-quality solutions than leading guidance-based approaches in the classic one-shot MAPF setting. Moreover, in the planning-and-execution MAPF setting, which emphasises anytime behaviour, our approach consistently outperforms the state-of-the-art solver PIE~\cite{zhang2024planning} across all evaluated configurations.

%% file: section/preliminary.tex
\section{Problem Definition}

In the (classic) Multi-Agent Path Finding (MAPF)~\cite{stern2019multi} problem, we model the environment as a graph $G=(V,E)$, where vertices $V$ represent locations and edges $E$ represent feasible moves between locations. We consider a set of $n$ agents $A=\{a_1,\dots,a_n\}$. Each agent $a_i$ is given a start vertex $s_i\in V$ and a goal vertex $g_i\in V$. Time is discretised into steps $t=0,1,2,\dots$. At each time step, an agent may either \emph{wait} (remain at its current vertex) or \emph{move} to an adjacent vertex.
A (discrete-time) path for agent $a_i$ is a sequence of vertices $\pi_i=\langle v_i(0),v_i(1),\dots,v_i(T_i)\rangle$ such that $v_i(0)=s_i$ and for all $t\in\{0,\dots,T_i-1\}$,
either $v_i(t+1)=v_i(t)$ (wait) or $(v_i(t),v_i(t+1))\in E$ (move). The agent reaches its goal at time $T_i$ with $v_i(T_i)=g_i$, and is assumed to remain at $g_i$ thereafter.
A solution is a set of paths $\Pi=\{\pi_1,\dots,\pi_n\}$ that is \emph{collision-free}. In this paper, we consider 
(i) \textbf{Vertex conflicts}: no two agents occupy the same vertex at the same time,
and (ii) \textbf{Edge conflicts} (swap conflicts): no two agents traverse the same edge in opposite directions at the same time.

\paragraph{Sum-of-loss objective.}
In this paper, we focus on finding a feasible joint plan $\Pi$ that minimises \emph{Sum-of-loss} $\mathrm{SoL}(\Pi)$.
For a feasible joint plan $\Pi=\{\pi_i\}_{i=1}^n$, 
let $T=\max_i t_i$, SoL can be written as
\[
\mathrm{SoL}(\Pi)=\sum_{t=0}^{T-1}\left|\left\{ i \in \{1,\dots,n\} \ \middle|\ \pi_i(t)\neq g_i \right\}\right|.
\]
Intuitively, at each timestep we count how many agents are still not at their goals, and SoL sums this
quantity over time. This objective therefore rewards solutions that bring agents to their goals earlier,
capturing overall progress rather than only the last arrival time.

\section{Related Works}
\subsection{PIBT and LaCAM}

\paragraph{Priority Inheritance and Backtracking (PIBT)} \cite{okumura2022priority}  constructs successor configurations incrementally.
Let $Q_t = (v_1^t, \dots, v_n^t)$ denote the current configuration.
For each agent $a_i$, the candidate action set is
$
\mathcal{A}_i(Q_t) = \{v_i^t\} \cup \{u \mid (v_i^t, u) \in E\}.
$
Agents are processed according to a priority ordering.
For agent $a_i$, actions $u \in \mathcal{A}_i(Q_t)$ are ranked by an evaluation function
$
f_i(u)
$
and the highest-ranked feasible action is selected.
If the selected action is blocked by a lower-priority agent, priority inheritance and backtracking are applied recursively to resolve conflicts. The effectiveness of PIBT strongly depends on the evaluation function $f_i$, which in original approaches is based solely on individual shortest-path distances i.e., $f_i(u) = \mathrm{dist}(u, g_i)$ in order to guide agents moving toward goal configuration.

\paragraph{Lazy Constraint Addition MAPF (LaCAM)}
\cite{okumura2023lacam} performs a search over the space of joint configurations.
At the high level, each search node corresponds to a configuration $Q$ and stores an accumulated transition cost $g(Q)$ representing the cost of the partial joint plan from the initial configuration to $Q$.
The high-level search proceeds in a depth-first manner.

Successor configurations are generated at the low level using PIBT.
Given a configuration $Q$, PIBT produces a feasible successor configuration $Q'$ by selecting one action per agent according to the evaluation function $f_i$ and resolving conflicts via priority inheritance and backtracking.
LaCAM* generates exactly one successor per expansion.
If a previously generated successor $Q'$ is revisited from the same parent $Q$, LaCAM* lazily expands a constraint tree (attached to $Q$) that systematically iterates through valid moves for agents, assigning specific `who is where' constraints that force PIBT to generate alternative successors.
This mechanism enables systematic enumeration of all feasible successors without explicitly branching over the joint action space.

Let $\mathcal{Q}$ denote the set of all reachable configurations.
LaCAM* is complete because the depth-first traversal, together with lazy constraint addition, ensures that every $Q \in \mathcal{Q}$ reachable from the initial configuration is eventually expanded, unless terminated by a time limit.
To guarantee eventual convergence to an optimal solution, LaCAM* maintains and updates accumulated costs $g(Q)$ and parent pointers.
Whenever a lower-cost path to an already discovered configuration $Q$ is found, $g(Q)$ is updated and the improvement is propagated to descendant configurations via Dijkstra-style relaxation.
In addition, LaCAM* incorporates a \emph{swap} operator into the low-level generator to resolve local livelocks, further reducing search effort in dense instances.

\subsection{Traffic Optimisation Using Guide Paths}
PIBT and LaCAM/LaCAM* rely on evaluation functions based on individual shortest-path distances, which ignore inter-agent interactions and often lead to severe congestion.
\emph{Traffic Flow Optimisation}~\cite{chen2024traffic} addresses this limitation by precomputing time-independent \emph{guide paths} using a traffic-flow model that penalises vertex usage (vertex congestion) and opposing movements (contraflow congestion).
These guide paths are obtained through a Frank--Wolfe-style iterative optimisation procedure.
During execution, agents prioritise moves that align with their assigned guide paths, significantly improving throughput.

Similarly, the enhanced LaCAM~\cite{okumura2023engineering} integrates \emph{Space Utilisation Optimisation} (SUO)~\cite{SUO}, which iteratively replans individual paths in a precomputation phase to minimise overlap on a global usage heatmap, producing spatially dispersed routes.
These routes serve as a structural bias within the low-level PIBT generator, discouraging agents from entering crowded bottlenecks and reducing search effort in dense environments.

In both approaches, the core mechanism is to replace the original PIBT sorting function
$f_i(u) = \mathrm{dist}(u, g_i)$
with a traffic-aware evaluation function derived from precomputed guide paths, thereby biasing action selection toward less congested regions.

%% file: section/algorithm.tex
\section{Our Method}

While recent traffic optimisations enhance solution quality, they face two primary limitations. 
First, the iterative path optimisation introduces non-negligible initialization overhead, which delays the first action and negatively impacts anytime performance. 
Second, this optimisation is performed only once prior to the search. 
While the first solution is often significantly improved, subsequent search iterations tend to plateau, yielding only marginal or slow improvements.

To address these limitations, we propose a simple yet efficient approach called \emph{Lightweight Traffic Map} (LTM). 
The main idea of our approaches leverages the ability of PIBT as a low-level successor generator to rapidly generate feasible configurations. That is, instead of performing iterative path optimisation, we directly collect historical data generated by PIBT to construct a lightweight traffic map. This traffic map is updated dynamically during search and is used to guide PIBT.
We also modify the LaCAM* search to iteratively restart from selected configurations.

\begin{algorithm2e}[t]
\small
\setcounter{AlgoLine}{0}
\SetKwInput{Input}{Input}
\SetKwInput{Output}{Output}
\SetKwInput{Initialization}{Initialization}

\Input{MAPF instance $\mathcal{I}$; initial root node $N$; per-iteration budget $B$; termination condition $\mathcal{T}$.}
\Output{Best solution found $S_{\text{best}}$.}

\Initialization{
$\mathrm{LTM} \gets$ UniformCostMap;
$S_{\text{best}} \gets \emptyset$\;
}

\While(\tcp*[f]{anytime loop}){$\mathcal{T}$ is not satisfied \label{line:termination}}{
    $(S, H) \gets \mathrm{LaCAM*}(\mathcal{I}, N, \mathrm{LTM}, B);$\tcp*[f]{$H$: PIBT history during this run}\; \label{line:run_lacam}

    \If{$S \neq \emptyset$ \textbf{and} ($S_{\text{best}}=\emptyset$ \textbf{or} $\mathrm{SoL}(S) < \mathrm{SoL}(S_{\text{best}})$)}{
        $S_{\text{best}} \gets S$\; \label{line:update_best_soluton}
    }

    $\mathrm{LTM} \gets \mathrm{UpdateLTM}(\mathrm{LTM}, H)$;\tcp*[f]{update traffic map from history}\;  \label{line:update_ltm}
    $N \gets \mathrm{SelectRestartNode}(H);$\tcp*[f]{select restart node}\;
    \label{line:select_restart_node}
}

\Return{$S_{\text{best}}$}\; \label{line:return}

\caption{LaCAM* + LTM}
\label{algo:lacam_ltm}
\normalsize
\end{algorithm2e}










Algorithm \ref{algo:lacam_ltm} presents the overall framework of our modified LaCAM* with the proposed LTM. The algorithm maintains an anytime search loop that iteratively improves solution quality until a termination condition is met (line~\ref{line:termination}). At each iteration, a one-shot LaCAM is called with a bounded node or time budget and guided by the current LTM (line~\ref{line:run_lacam}).
At initialisation, LTM is initialised with uniform traversal costs identical to those of the original map (i.e., unit cost), such that the first bounded LaCAM* execution is equivalent to the original LaCAM*. The best solution found so far is initialised to null, and the search starts from an initial restart node.
During the first LaCAM run, PIBT is used as the low-level configuration generator, and its execution history is recorded. 

After LaCAM returns, first, 
the resulting solution is then compared against the current best solution, and the best one is retained (line~\ref{line:update_best_soluton}).
Then, the algorithm directly updates LTM using the collected PIBT history (line~\ref{line:update_ltm}). The details of maintaining and updating LTM are presented in the next section. Based on the updated LTM and the accumulated search history, a new restart node (or configuration) is selected to initialise the next iteration (line~\ref{line:select_restart_node}). This restart mechanism enables LaCAM* to iteratively explore alternative regions of the search space while continuously refining its traffic guidance. Finally, once the termination condition is satisfied, the algorithm returns the current best solution (line~\ref{line:return}).

\subsection{Updating the Lightweight Traffic Map}
To ensure efficiency, we intentionally keep the design of the LTM simple. The LTM closely follows the structure of the original MAPF graph that defines traversability. However, instead of using uniform costs on undirected edges, the LTM assigns adaptive weights to directed edges to capture traffic information observed during the search.

\begin{defn}\textbf{(Lightweight Traffic Map)}
Given a MAPF instance represented by a graph $G=(V,E)$, the Lightweight Traffic Map is defined as a directed weighted graph $G_{\text{LTM}}=(V_L,E_L,w_L)$, where $V_L = V$. For each undirected edge $(v_i,v_j) \in E$, the edge set $E_L$ contains two directed edges $(v_i,v_j)$ and $(v_j,v_i)$. The weight function $w_L$ assigns a non-negative cost to each directed edge, with weights normalized to lie within a bounded interval $w_{LB} \le w_L(e) \le w_{UP}$.
\end{defn}

The key idea of LTM is to leverage traffic information (actions and conflict-induced blockage) observed from PIBT executions to update edge weights dynamically. The edge weight reflects its observed traffic condition: edges with higher weights indicate heavier traffic and are therefore less preferred during traversal. By incorporating this information, LTM implicitly penalises congested transitions and encourages agents to explore alternative routes.

To prevent excessive penalization that could completely block certain transitions, all edge weights are normalized within fixed lower and upper bounds $[w_{LB}, w_{UP}]$. This bounded design ensures that no edge becomes prohibitively expensive and preserves the connectivity of the graph. In our experiments, we set $[w_{LB}, w_{UP}] = [0,10]$, although this range can be easily tuned for different maps or domains.
Note that, our LTM shares conceptual similarities with guidance graphs~\cite{OnlineGGO}, but differs in several important aspects. First, we do not explicitly model self-loop edges corresponding to wait actions. Instead, congestion at a vertex is implicitly propagated to its adjacent edges. Second, unlike guidance graphs that rely on expensive offline random instance sampling and training, often requiring hours or days to construct, our LTM directly encodes traffic information gathered online from previous PIBT executions. As a result, LTM can be built and updated efficiently during the search.
Next, we describe how the LTM is updated using historical data collected from PIBT.

\subsubsection*{Collecting Traffic Information from PIBT}

Recall that LaCAM employs the low-level configuration generator PIBT to construct the next configuration. Given a configuration $Q_{\text{from}}$, PIBT determines the next location for each agent $a_i$ by sorting the candidate vertices
$\text{neigh}(Q_{\text{from}}[a_i]) \cup \{Q_{\text{from}}[a_i]\}$
according to an evaluation function $f$. Initially, this function is defined as
$f(v_i) = \text{dist}(v_i, g_i)$
where $\text{dist}(v_i, g_i)$ denotes the shortest-path distance from vertex $v_i$ to the goal $g_i$ of agent $a_i$.
During each iteration of LaCAM, we collect two types of traffic-related data from PIBT transitions between configurations $Q_{\text{from}}$ and $Q_{\text{to}}$.

\begin{itemize}
    \item \textbf{Committed Actions.}
    For each agent $a_i$, we record the committed action $(Q_{\text{from}}[a_i], Q_{\text{to}}[a_i])$. This action corresponds to the actual movement chosen by PIBT and is used to increase the traffic cost of the corresponding edge. This is motivated by the observation that PIBT, when guided primarily by shortest-path distances, often directs multiple agents toward the same regions, leading to congestion. Penalizing frequently committed actions helps mitigate such traffic accumulation.

    \item \textbf{Blocked Actions.}
    For each agent $a_i$, PIBT may fail to select the highest-ranked action according to $f$ due to conflicts with higher-priority agents, which also indicate potential congestion. We therefore record these as blocked actions of the form $(Q_{\text{from}}[a_i], v_i)$, where
   $v_i \in \text{neigh}(Q_{\text{from}}[a_i]) \cup \{Q_{\text{from}}[a_i]\}$ and $f(u) < f(Q_{\text{to}}[a_i]).$
    
\end{itemize}

For each recorded action $(v_i, v_j)$ during a LaCAM iteration, we increment the corresponding directed edge cost by 1. Since recorded actions may include wait actions (i.e., $v_i = v_j$), we propagate the increased cost to all outgoing edges from $v_i$, 
i.e., $\forall u \in \mathrm{neigh}(v_i), w_L(v_i, u) = w_L(v_i, u) + 1$. This propagation allows vertex congestion to influence adjacent transitions without explicitly modeling self-loop edges. In addition, when an agent has already reached its goal, we ignore any subsequent wait actions and do not increase the corresponding traffic values. This design prevents goal locations from being artificially penalized due to agents waiting for others to finish, which is particularly important in instances with large makespan where prolonged waiting at goals is common.

To ensure correct normalization, we maintain a separate accumulator that records the raw traffic counts. After each update, the accumulated values are normalized into the predefined range $[w_{LB}, w_{UP}]$ to obtain the final LTM weights. During preliminary experiments, we explored more complex handcrafted update functions, but they did not outperform this simple additive scheme. While more sophisticated update rules may exist, designing optimal traffic update functions is beyond the scope of this paper.

\subsection{Modifying LaCAM* with LTM}
To efficiently use LTM in LaCAM*, we made two modifications to the original algorithm: (i) integrating LTM into the low-level successor generator PIBT used in each one-shot LaCAM; and (ii) modifying the anytime behaviour of LaCAM* to have more frequent restarts. 

\subsubsection*{Integrating LTM into PIBT}
Given an updated LTM, integrating it into PIBT is straightforward. Recall that PIBT relies on two key factors that influence the quality of the generated configurations: (i) the evaluation function $f(v_i)$, which determines the action ordering for each agent, and
(ii) the priority ordering among agents, which determines conflict resolution when multiple agents compete for the same vertex. We modify PIBT to incorporate LTM in the following ways:

\begin{itemize}
\item \textbf{Evaluation function.}
We replace the original evaluation function $f(v_i) = \text{dist}(v_i, g_i)$, computed on a uniform-cost graph, with the shortest-path distance computed on the weighted LTM. We further optimise the implementation to compute the distance on the dynamically updated weighted graph with a backward continuous A* search using the Manhattan heuristic to avoid precomputation overhead. Since the LTM is updated frequently, the backward A* search is restarted from scratch after each update. Compared to the original implementation, this modification introduces only negligible overhead.

\item \textbf{Agent priority.}
In the original one-shot LaCAM*, agent priorities at the root node are assigned based on the shortest distance $\text{dist}(s_i, g_i)$, and subsequent PIBT executions inherit priorities from parent nodes. We modify this priority assignment to use distances computed on the updated LTM.

\item \textbf{Other LaCAM* Advances.} We also update distance estimates used in advanced PIBT optimisations, such as the push-and-swap operator used in LaCAM*, to rely on shortest-path distances computed on the updated LTM.
\end{itemize}


\subsubsection{Modifying LaCAM* with frequent restarts}
The original LaCAM* operates anytime in a depth-first search manner: it always generates one successor with PIBT and expands on that successor; once the goal configuration is met, it backtracks to the parent of the goal node to generate another successor. This iteration runs until all configurations are seen or pruned. In other words, it can be seen as iteratively running one-shot LaCAM*, growing the search tree and restarting at the last generated node, which is also shown in Algorithm~\ref{algo:lacam_ltm}. This process could have potential bottlenecks from (i) one LaCAM* iteration struggles to find the goal within a short time; (ii) the backtrack logic could lead to similar or worse solutions, such as agents taking local detours, which slows down the anytime performance.
Therefore, we further modify the search to restart more frequently in two ways. First we introduce three early termination conditions to further restrict search effort and improve efficiency. 

\begin{itemize}
\item
The current iteration is terminated as soon as a goal configuration is found. Alternatively, if the search encounters an existed configuration and successfully obtain a better solution, the iteration is also terminated. This design preserves LaCAM’s anytime behaviour by immediately returning improved solutions.

\item 
We also terminate the iteration when the evaluation value of a search node, $f(n)=g(n)+h(n)$, exceeds the cost of the current best solution $S_{\text{best}}$. This prevents collecting traffic information from search regions that are unlikely to contribute to solution improvement.

\item  
Each LaCAM* iteration is also constrained by a predefined node  budgets. This condition is motivated by the observation that LaCAM* may generate a large number of low-level constraint tree nodes to resolve complex conflicts when PIBT fails to produce a feasible configuration, which can dominate the search time.

\end{itemize}

Next, we describe how restart nodes are selected and how PIBT is modified to incorporate guidance from the LTM.
In principle, any node recorded in the current LaCAM* search tree can be selected as a restart node. However, selecting a restart node uniformly at random is unlikely to yield better solutions. A common and effective strategy is to restart from nodes along search branches that lead to a solution, as these nodes are more likely to guide the search toward improved solutions. The specific restart strategy can be adapted to different application scenarios; in next section, we present concrete strategies for both one-shot MAPF and planning-and-execution settings.


\section{Putting it Together}
In this section, we discuss how the proposed algorithm, LaCAM*+LTM, can be adapted to two important application settings:
(i) one-shot MAPF, where the algorithm is given a fixed time budget to solve a MAPF instance in a single run and returns the best solution found within that budget; and
(ii) planning-and-execution MAPF, where the solver is repeatedly given a short time budget and is required to immediately commit a number of actions for execution.

\subsection{One-shot MAPF}
Adapting LaCAM*+LTM to the one-shot (classic) MAPF setting is straightforward. To preserve LaCAM’s superior performance in finding an initial solution, we do not impose any node budget in the first iteration; consequently, the first iteration proceeds identically to the original LaCAM*. From the second iteration onward, we limit each LaCAM* run by a node budget set to ten times the current makespan (i.e., the depth of the search tree). This bound prevents the search from becoming trapped in the exponential growth of the low-level constraint tree when resolving complex conflicts.

Regarding restart node selection, we observe that, given sufficient runtime, restarting from the root node almost always leads to better convergence and higher-quality solutions. However, for time-constrained applications, this strategy can be adapted by selecting restart nodes from the configuration that are closer to a goal configuration.

\subsection{Planning and Execution MAPF}
Unlike one-shot MAPF, the planning-and-execution setting models a more practical scenario in which robots cannot wait indefinitely to compute a complete plan and must begin executing actions while planning continues. We adopt the standard planning-and-execution model~\cite{zhang2024planning}, parameterized by the execution time $E$ and the commitment horizon $X$. Each action requires $E$ seconds to execute, and the agent must commit to executing $X$ actions at a time. Consequently, execution opens a planning window of duration $E \times X$, during which the planner attempts to improve the solution from the current committed configuration toward the goal. At the initial step, a planning window of length $E \times X$ is available. If no solution is found within this window, the agent commits to $X$ wait actions and continues planning in the next window. This process repeats until a solution is found and execution can proceed.

To adapt LaCAM*+LTM to this setting, we align the termination condition of each LaCAM*+LTM execution with the planning window of length $E \times X$. LaCAM*+LTM is repeatedly executed until all agents reach their goals. Each execution follows the same one-shot MAPF setting, except that the search is not restarted from scratch: the global search tree is preserved and reused across executions. At the beginning of each planning window, we select the restart node $N_{\text{cur}}$ corresponding to the agents’ current configuration and restrict LaCAM* to expand only the subtree rooted at $N_{\text{cur}}$ using depth-first search. This design allows the search to focus on refining future actions while respecting already committed executions.
Between successive executions of LaCAM*+LTM, we apply the following modifications:
\begin{itemize}
\item Instead of returning the complete best solution $S_{\text{best}}$, we backtrack the solution and return only the next $X$ actions starting from $N_{\text{cur}}$ for execution.

\item After updating $N_{\text{cur}}$, we discard historical PIBT data collected prior to $N_{\text{cur}}$, remove the associated traffic increments, and re-normalize the LTM. This prevents outdated traffic information from biasing future planning.
\end{itemize}
Finally, if LaCAM*+LTM fails to find a solution within the initial planning window, the search is paused at its termination point. In the subsequent planning window, LaCAM*+LTM resumes from the previously paused search state rather than restarting from the root. 


%% file: section/experiment.tex
\section{Experiments}
We implement our proposed LaCAM*+LTM in C++. All experiments were performed on a Linux workstation running Ubuntu 24.04.2 LTS, equipped with an Intel Xeon W-2135 CPU (6 cores, 12 threads, 3.70 GHz) and 125 GB RAM. For reproducibility, the source code are available on repository~\footnote{https://github.com/will-published-after-acceptance.com}

\begin{figure*}[t]
\scriptsize
\begin{tabular}{@{~}llll@{~}}
\begin{minipage}{.24\linewidth}
  \centering
\includegraphics[scale=0.21]{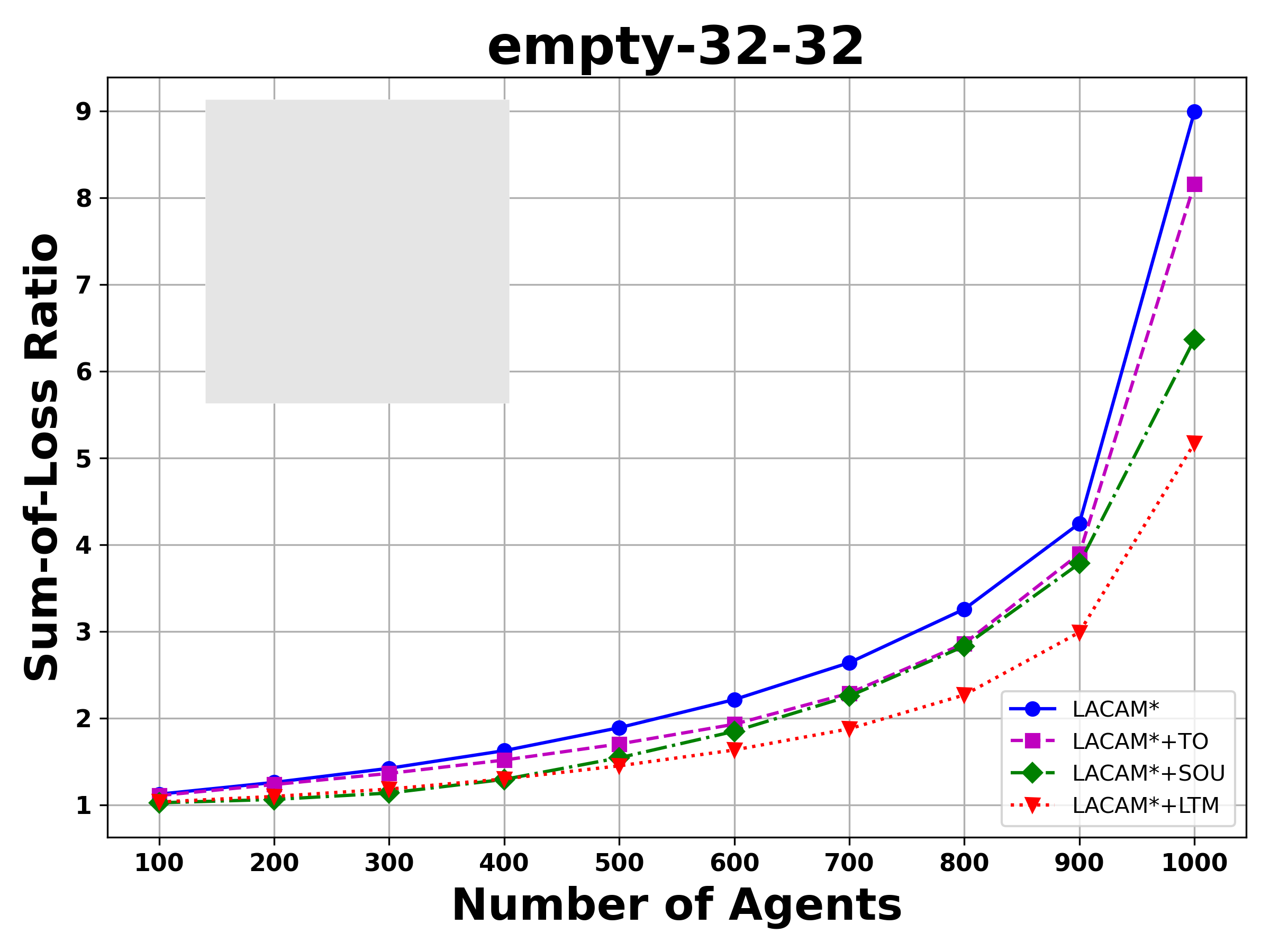}
\end{minipage}&

\begin{minipage}{.24\linewidth}
\centering
\includegraphics[scale=0.21]{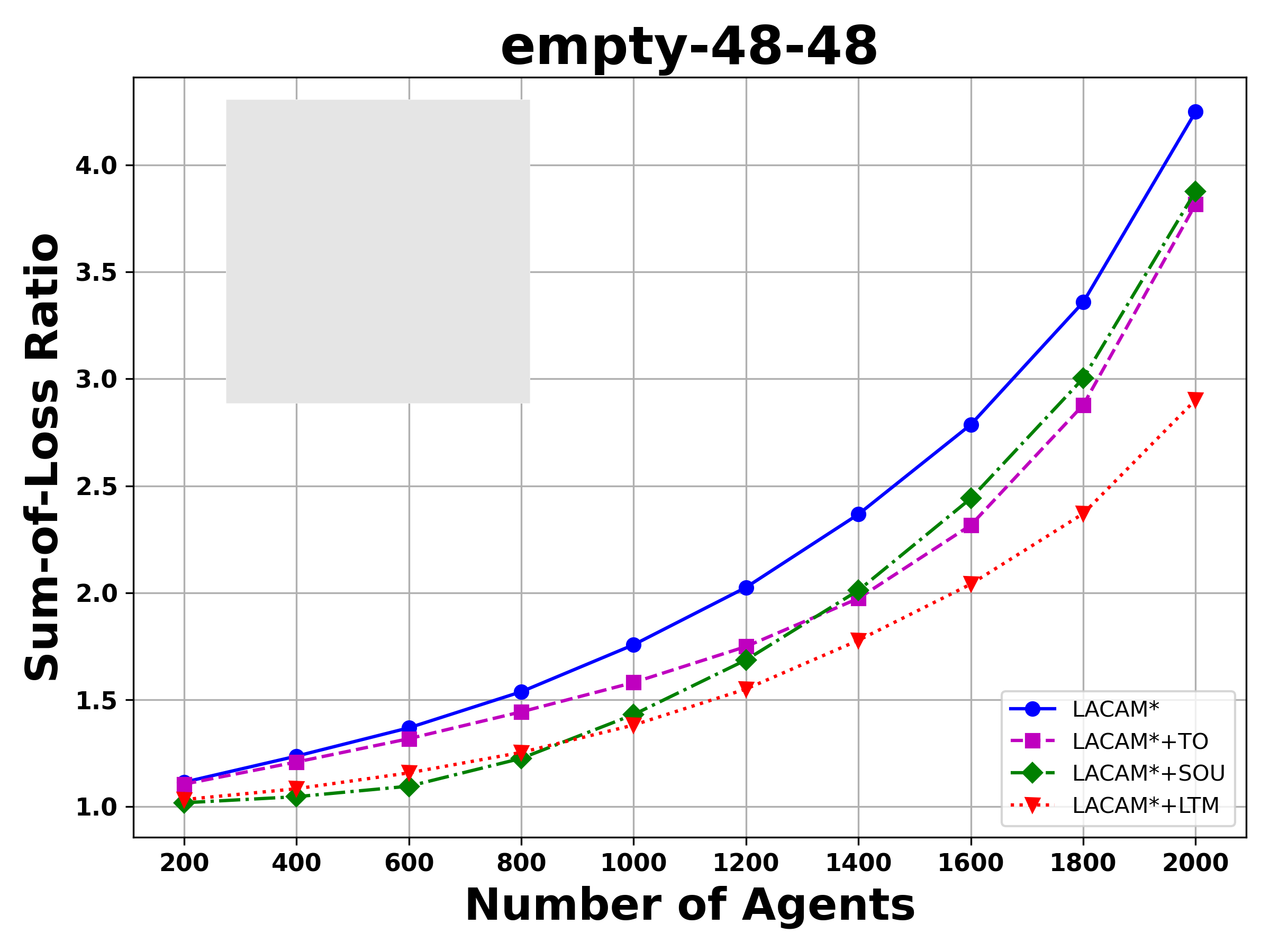}
\end{minipage}

\begin{minipage}{.24\linewidth}
\centering
\includegraphics[scale=0.21]{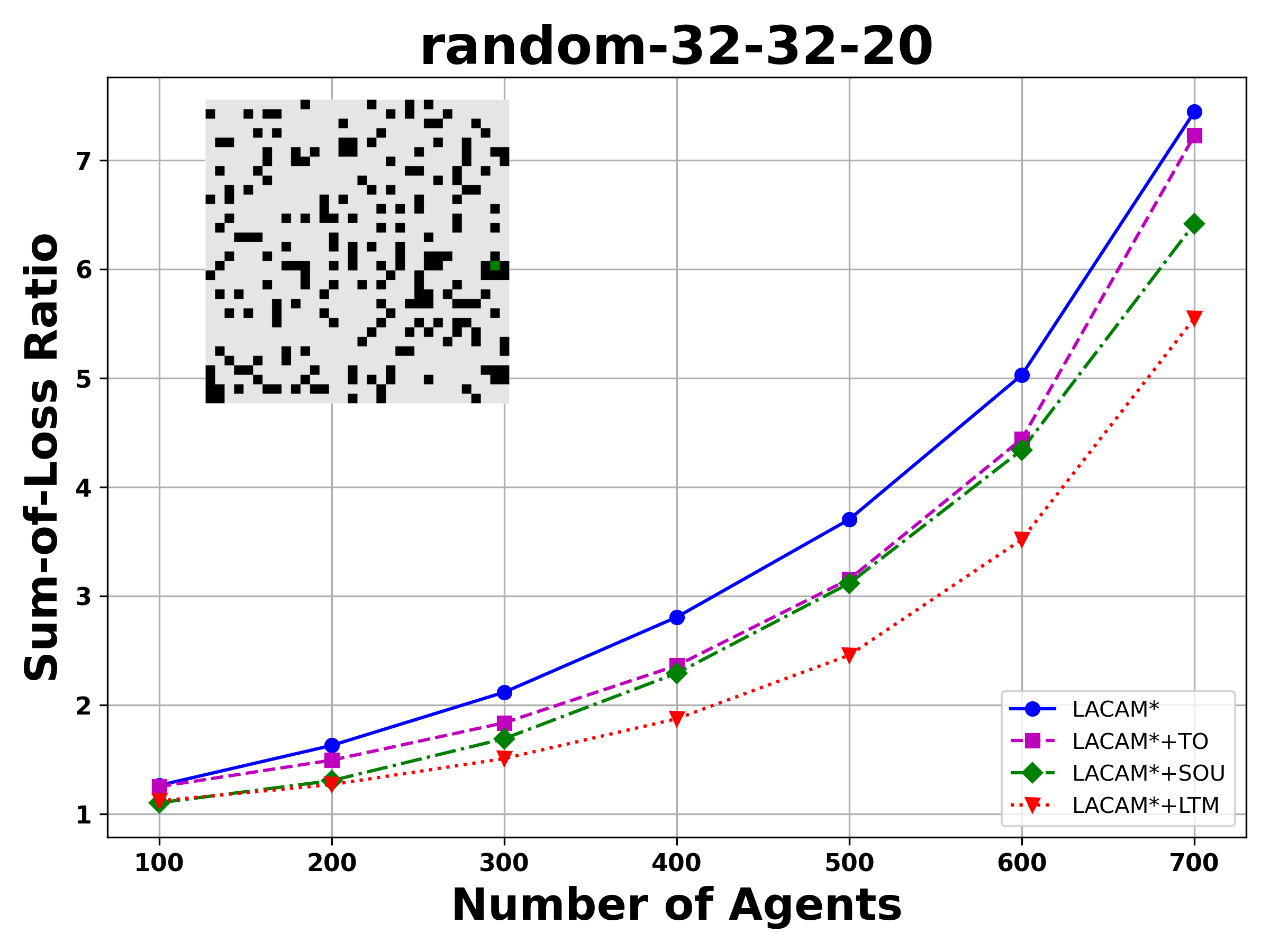}
\end{minipage}

\begin{minipage}{.24\linewidth}
\centering
\includegraphics[scale=0.21]{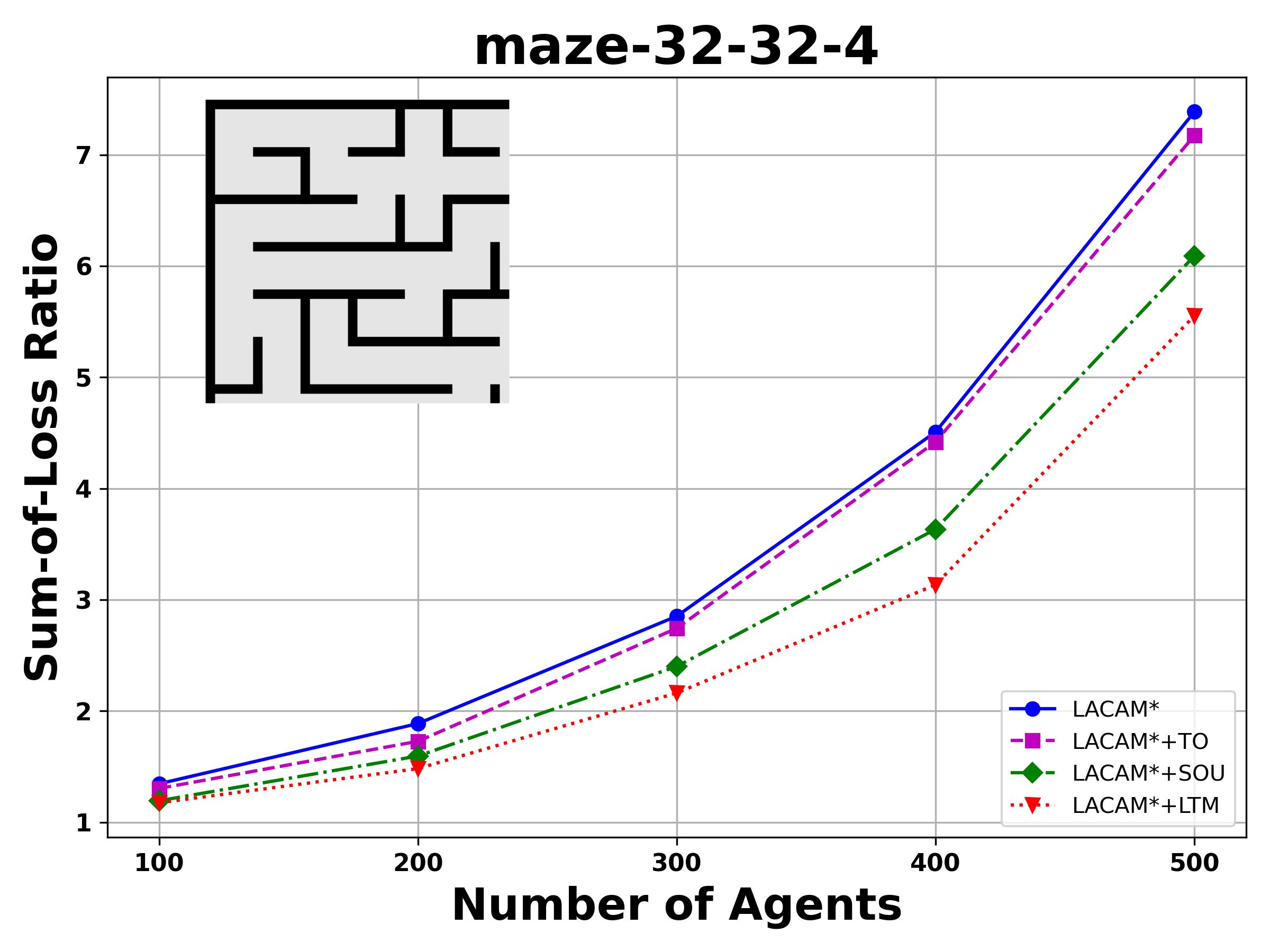}
\end{minipage}
\\
\begin{minipage}{.24\linewidth}
  \centering
\includegraphics[scale=0.21]{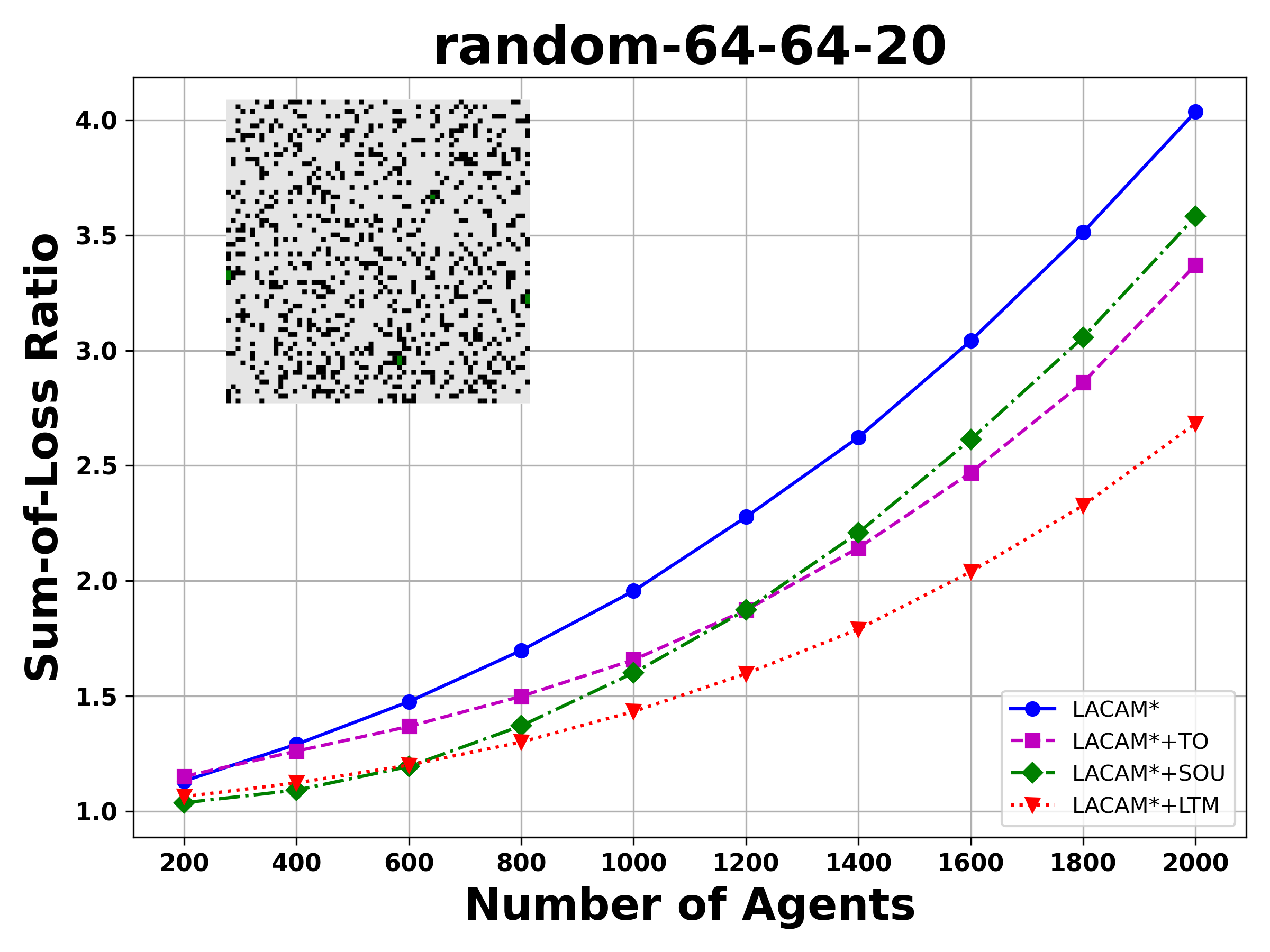}
\end{minipage}&

\begin{minipage}{.24\linewidth}
\centering
\includegraphics[scale=0.21]{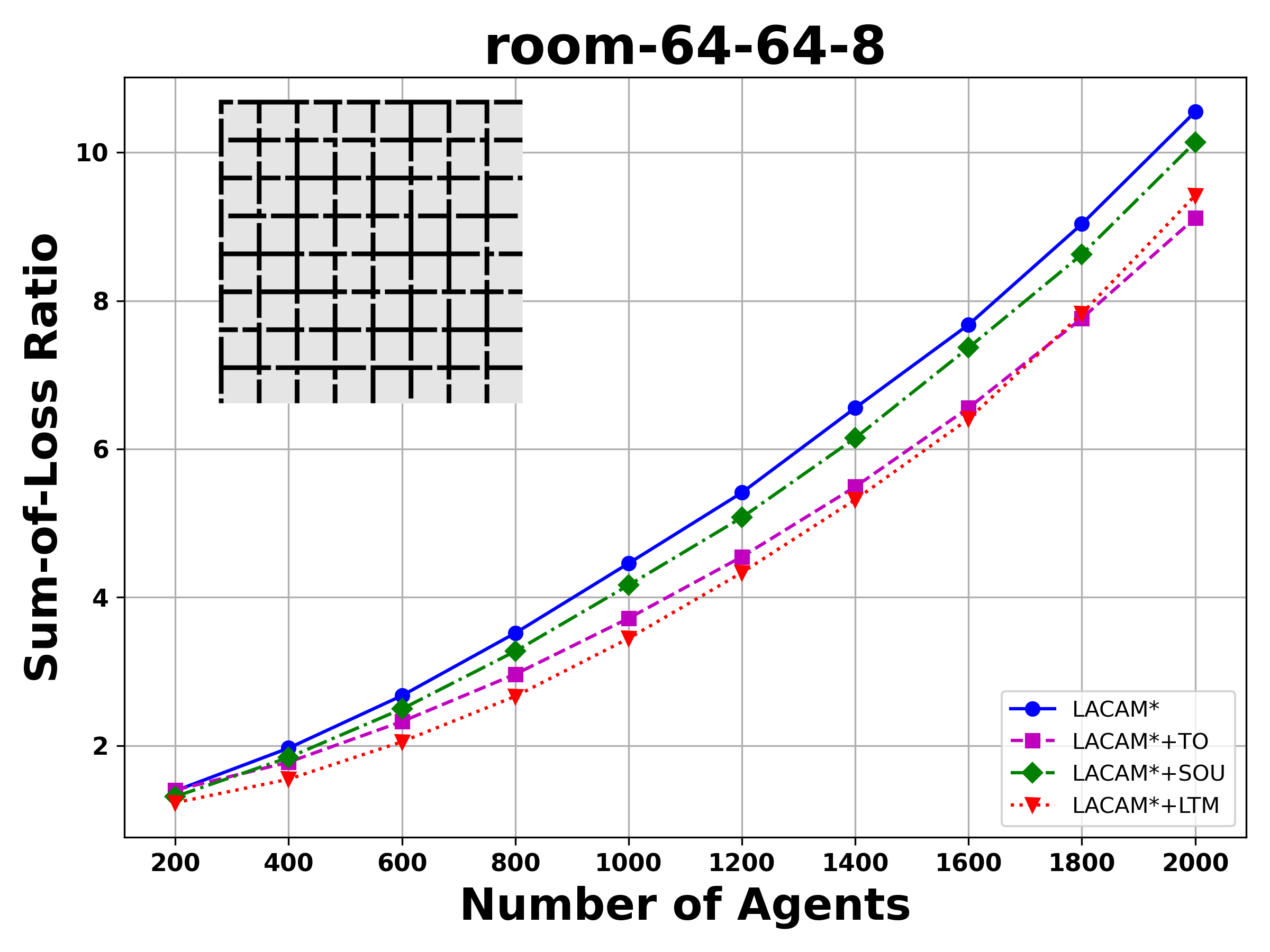}
\end{minipage}

\begin{minipage}{.24\linewidth}
\centering
\includegraphics[scale=0.21]{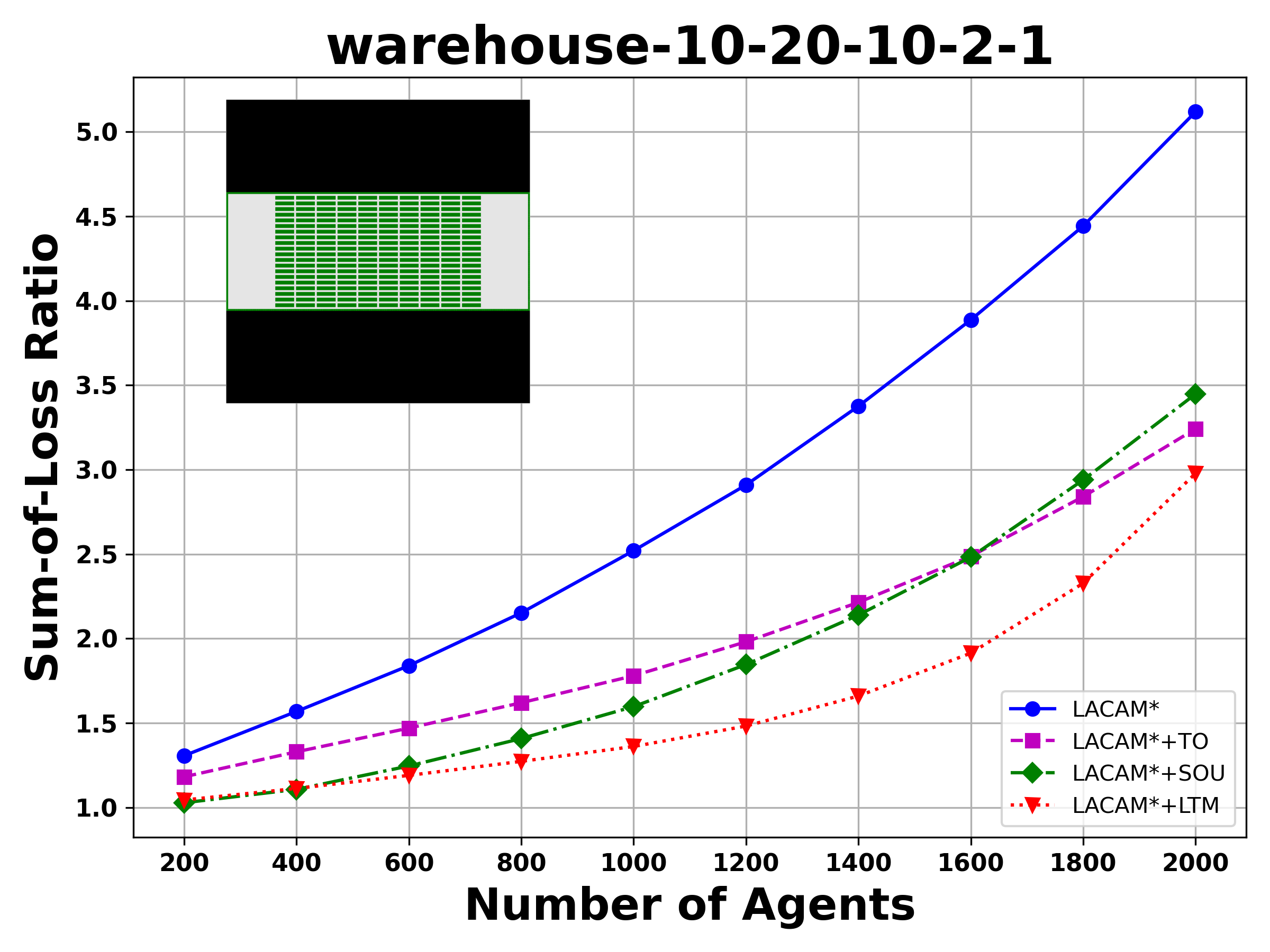}
\end{minipage}

\begin{minipage}{.24\linewidth}
\centering
\includegraphics[scale=0.21]{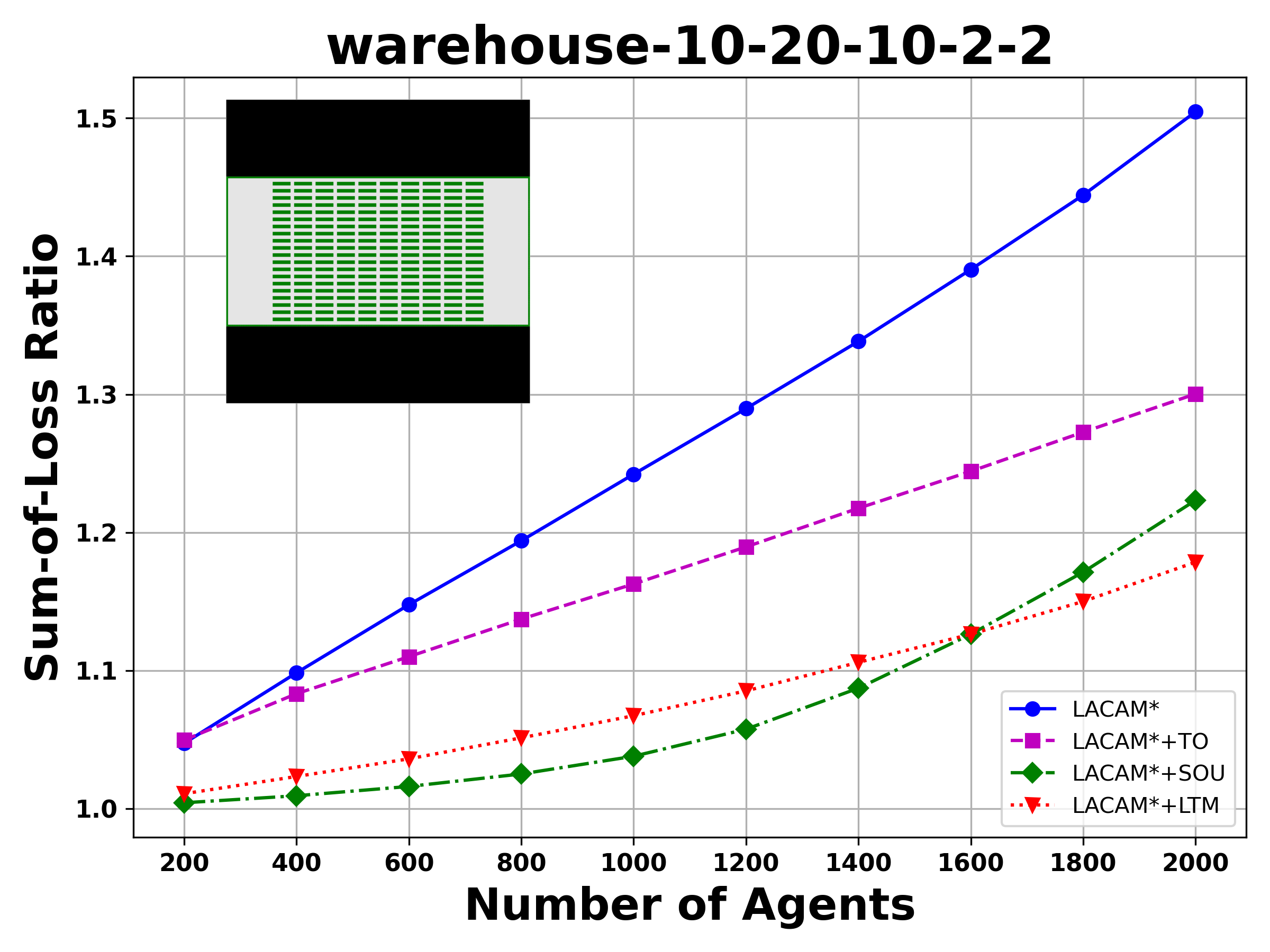}
\end{minipage}
\end{tabular}

\caption{Comparison of solution quality on eight grid-based MAPF benchmarks under the one-shot setting. Each plot shows the sum of loss ratio as a function of the number of agents for LaCAM*, LaCAM*+TO, LaCAM*+SUO, and the proposed LaCAM*+LTM, averaged over 25 random instances with a 30-second time limit. Lower values indicate better solution quality.}
\label{fig::solution_quality}
\end{figure*}









\begin{figure}[t!]
\scriptsize
\begin{tabular}{@{~}rr@{~}}
\begin{minipage}{.48\linewidth}
\centering
\includegraphics[width=\linewidth]{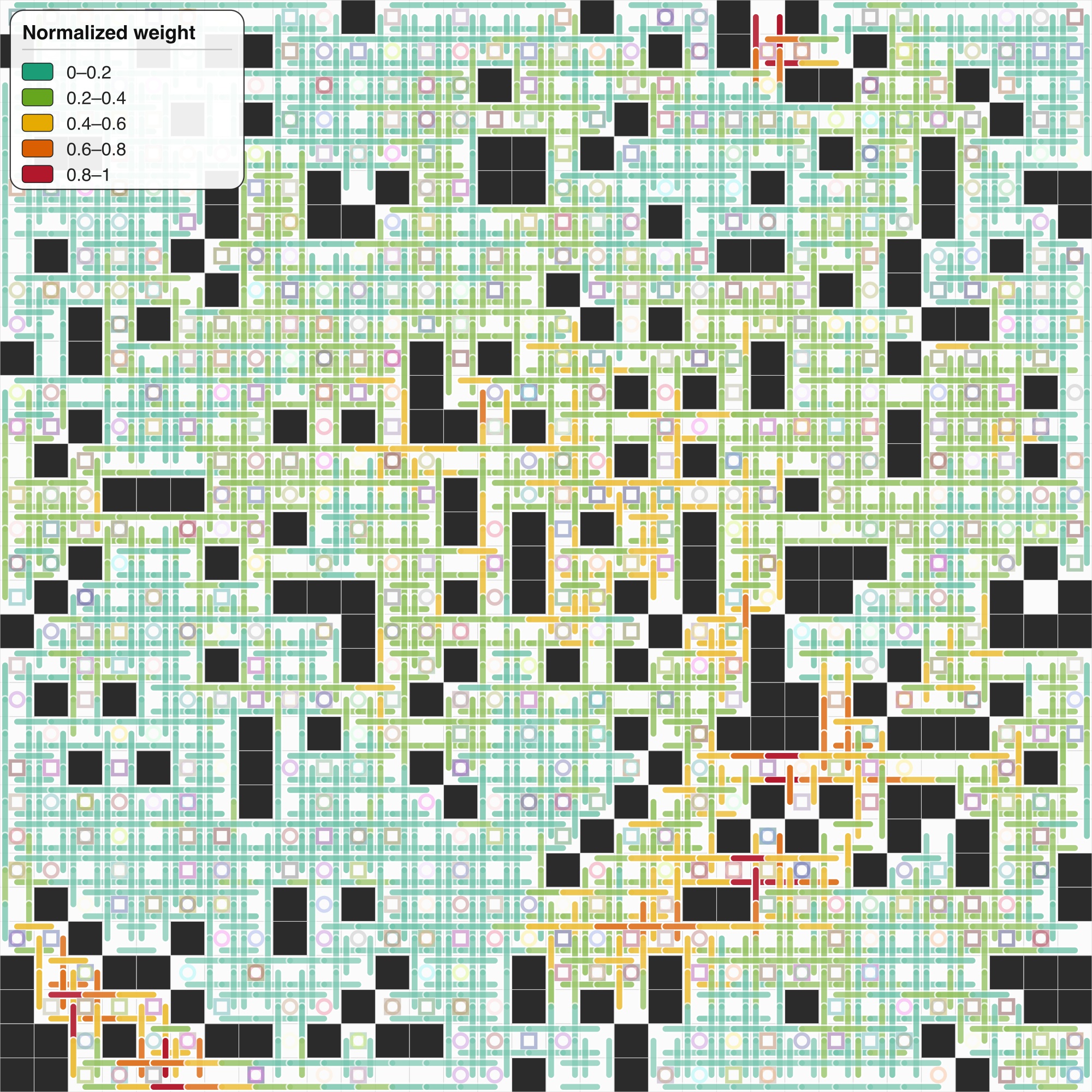}
\end{minipage}&

\begin{minipage}{.48\linewidth}
\centering
\includegraphics[width=\linewidth]{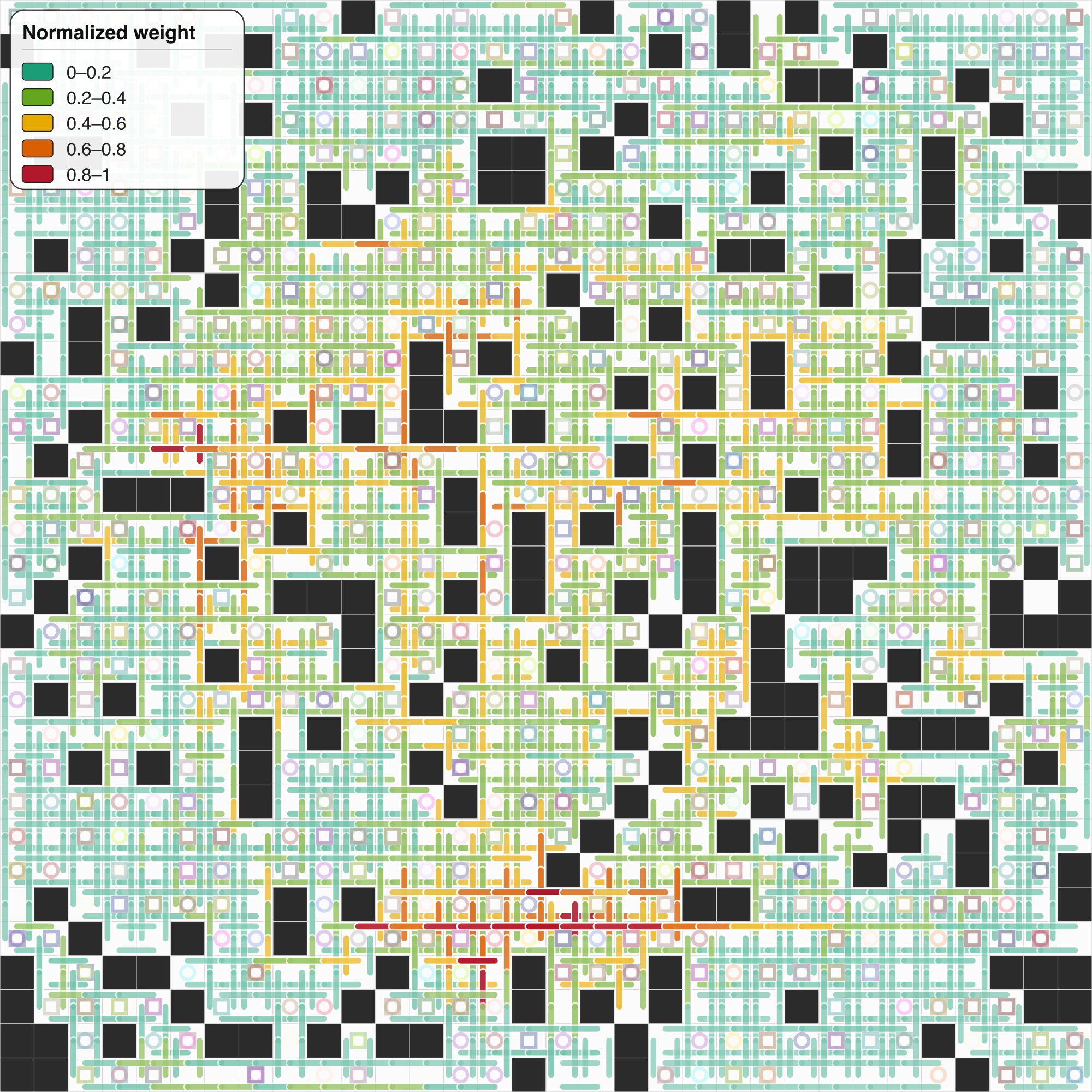}
\end{minipage}\\

\end{tabular}

\caption{ Visualization of the Lightweight Traffic Map (LTM) on the random-32-32-20 map with 400 agents. Circles denote start locations and squares denote goal locations. Edge colors indicate normalized traffic intensity. The LTM is shown after early iterations (left) and after 30 seconds of search (right).}
\label{fig::traffic_map}
\end{figure}

\begin{figure}[t!]
\centering
\scriptsize
\includegraphics[width=\linewidth]{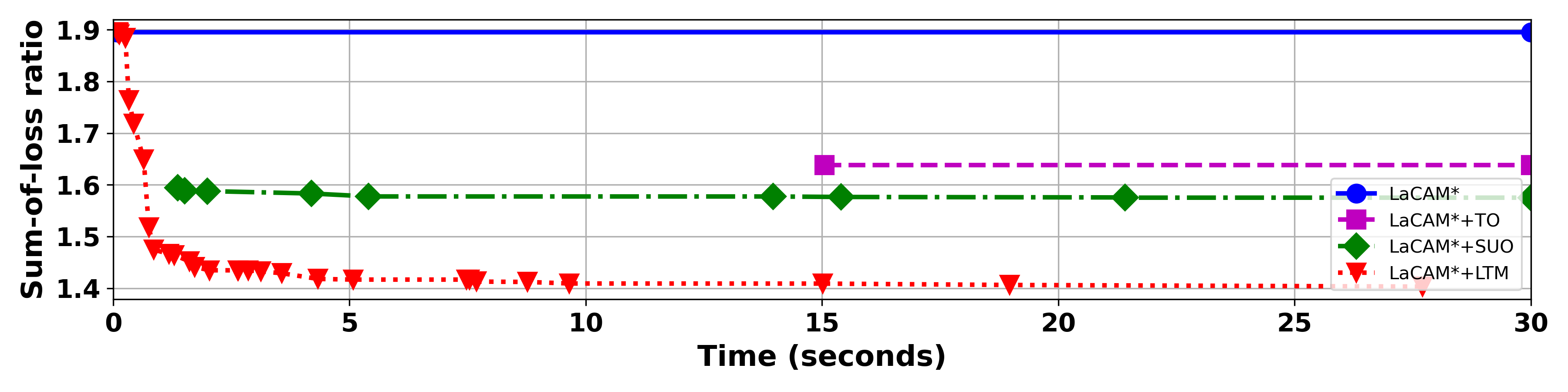}
\caption{Coverage plot over a 30-second runtime for 1000 agents on the random-64-64-20.map, comparing LaCAM, LaCAM+TO, LaCAM*+SUO, and the proposed LaCAM*+LTM.}
\label{fig::coverage_plot_64x64_1000}
\end{figure}

\begin{figure*}[t!]
\scriptsize
\begin{tabular}{@{~}llll@{~}}
\begin{minipage}{.24\linewidth}
  \centering
\includegraphics[scale=0.21]{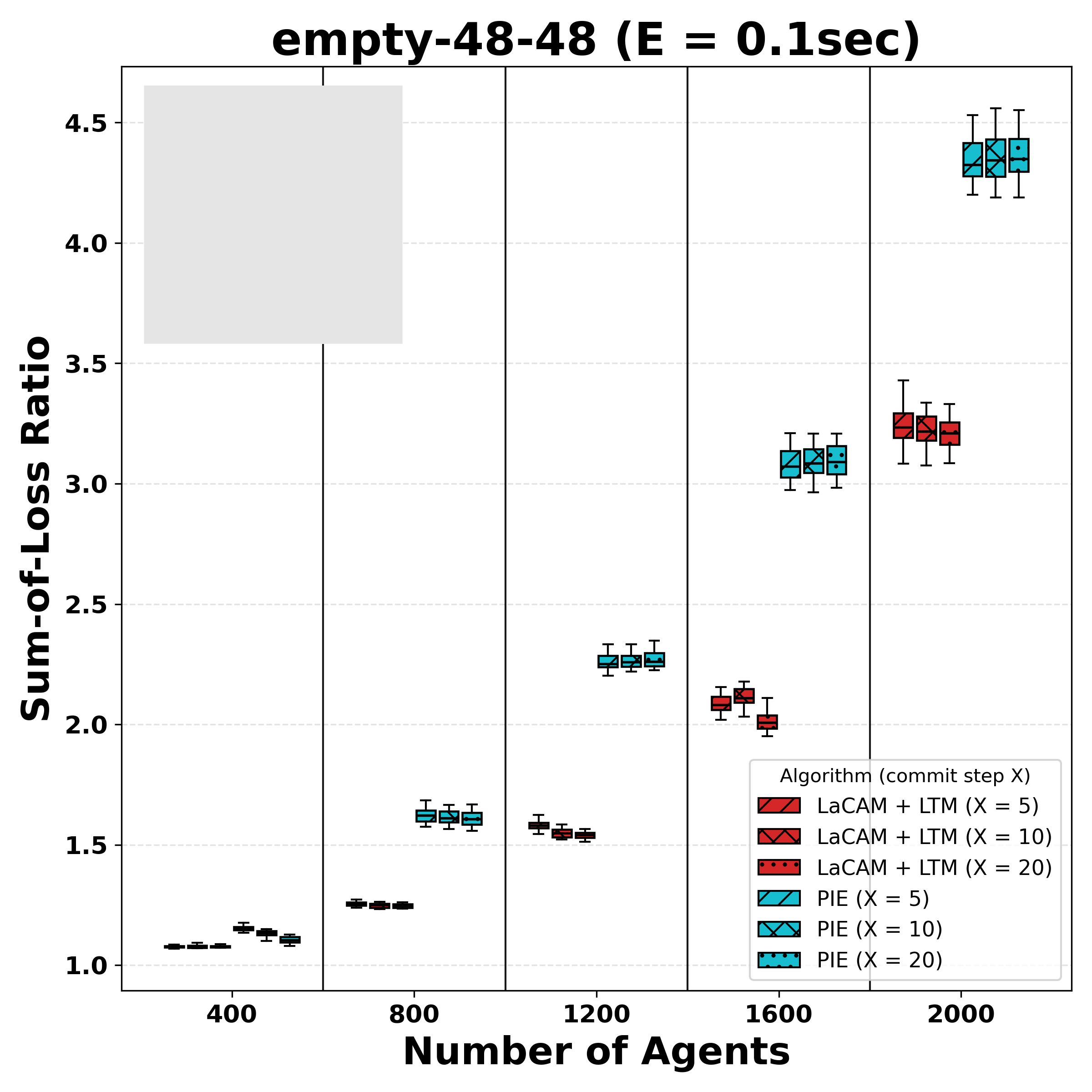}
\end{minipage}&

\hfill

\begin{minipage}{.24\linewidth}
\centering
\includegraphics[scale=0.21]{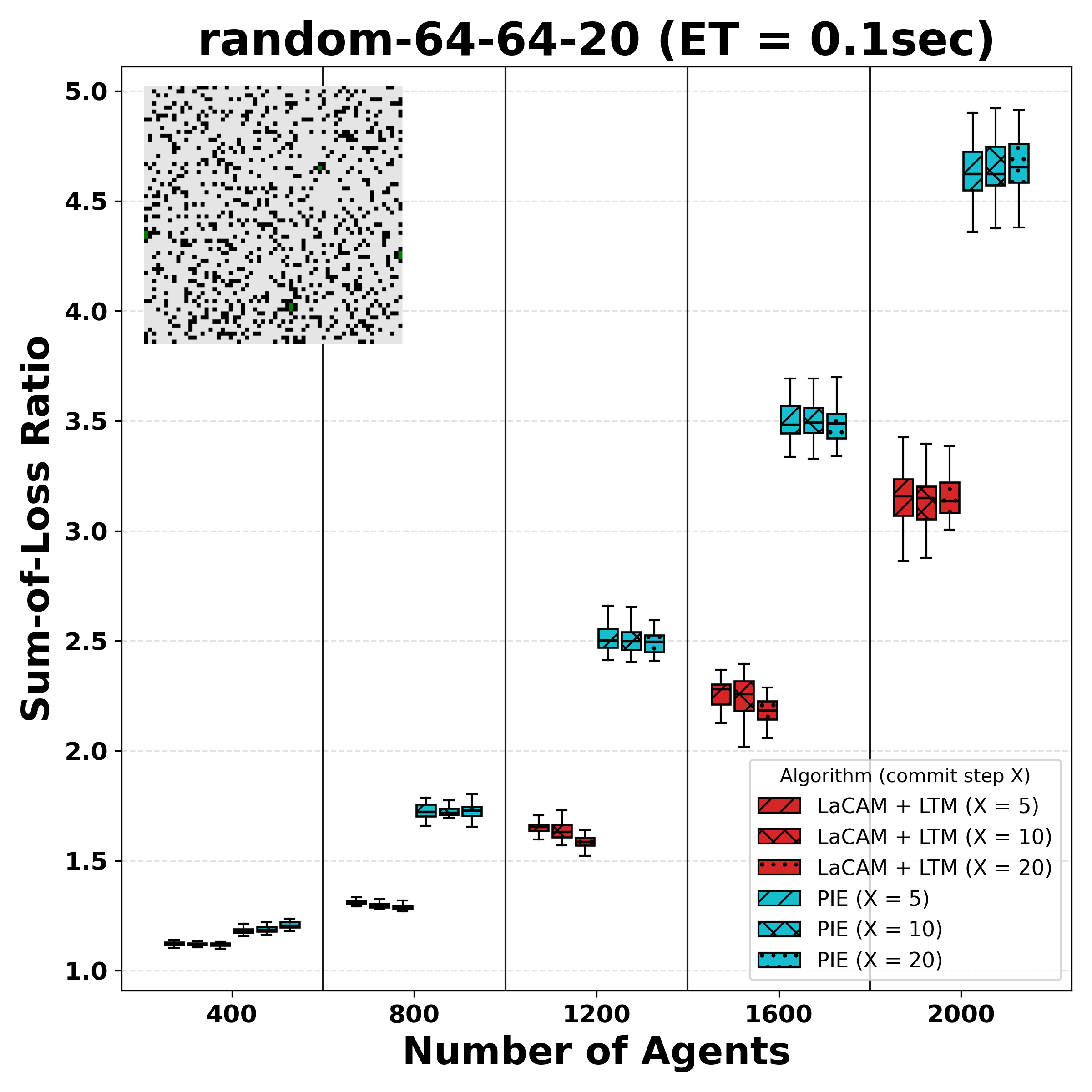}
\end{minipage}

\hfill

\begin{minipage}{.24\linewidth}
\centering
\includegraphics[scale=0.21]{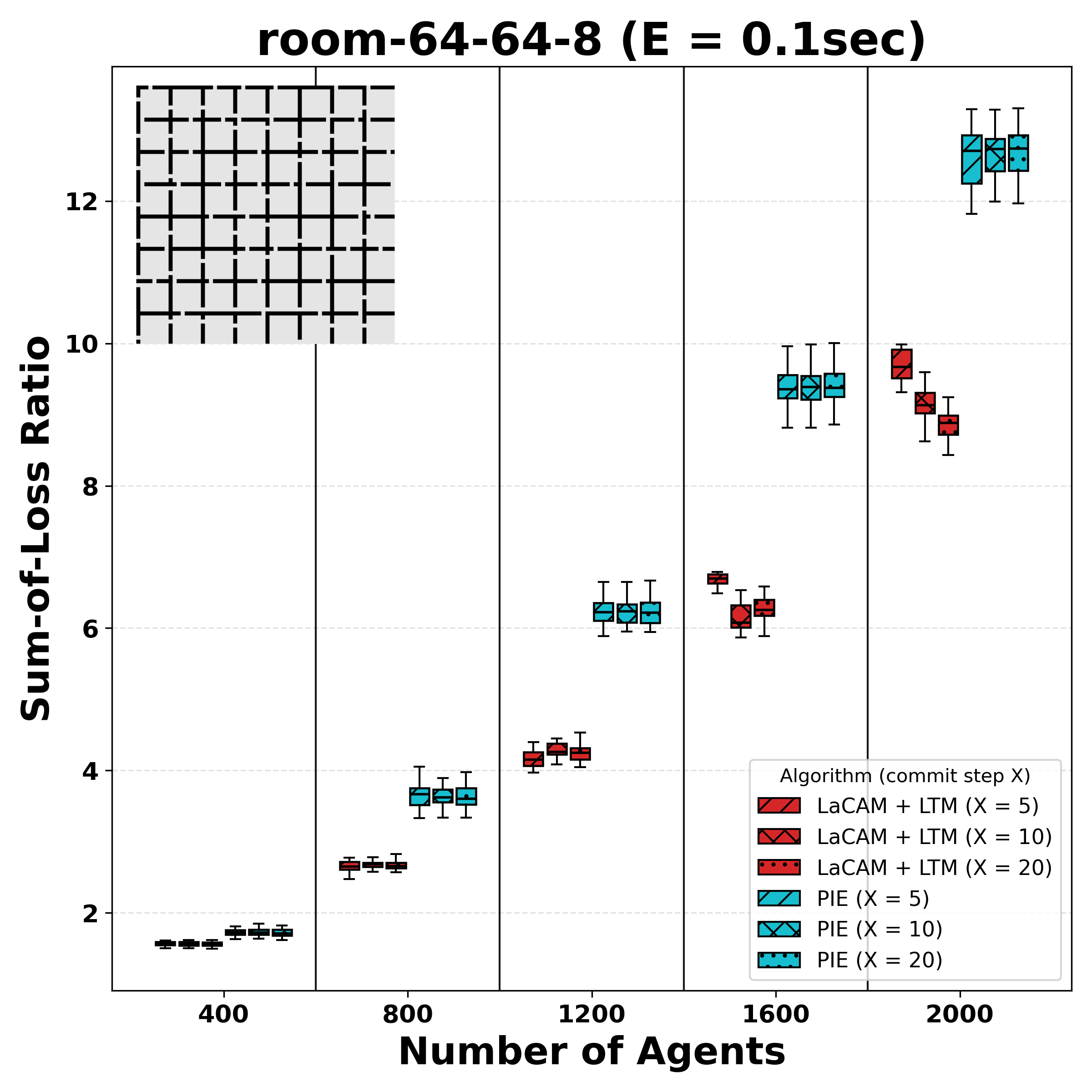}
\end{minipage}

\hfill

\begin{minipage}{.24\linewidth}
\centering
\includegraphics[scale=0.21]{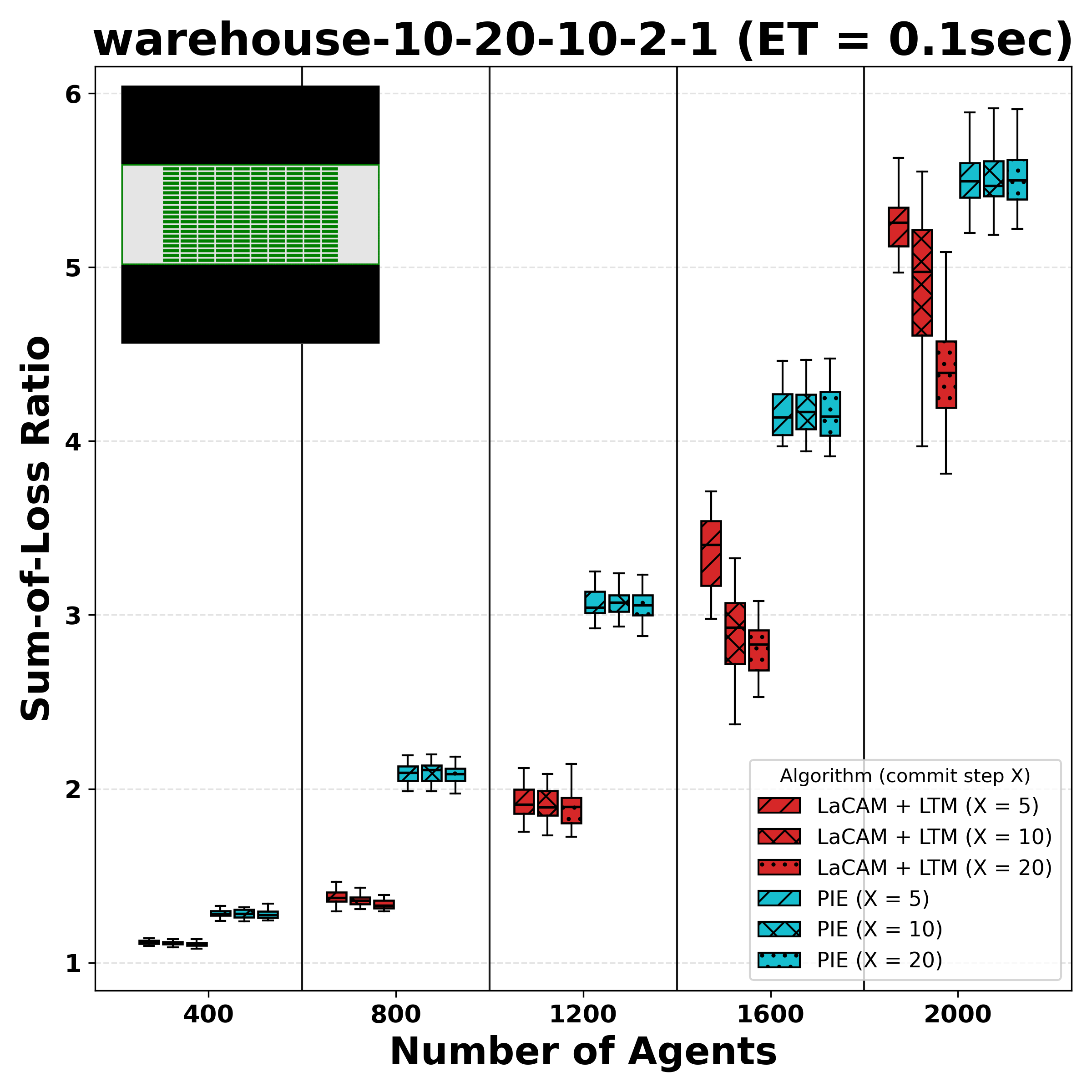}
\end{minipage}
\\
\begin{minipage}{.24\linewidth}
  \centering
\includegraphics[scale=0.21]{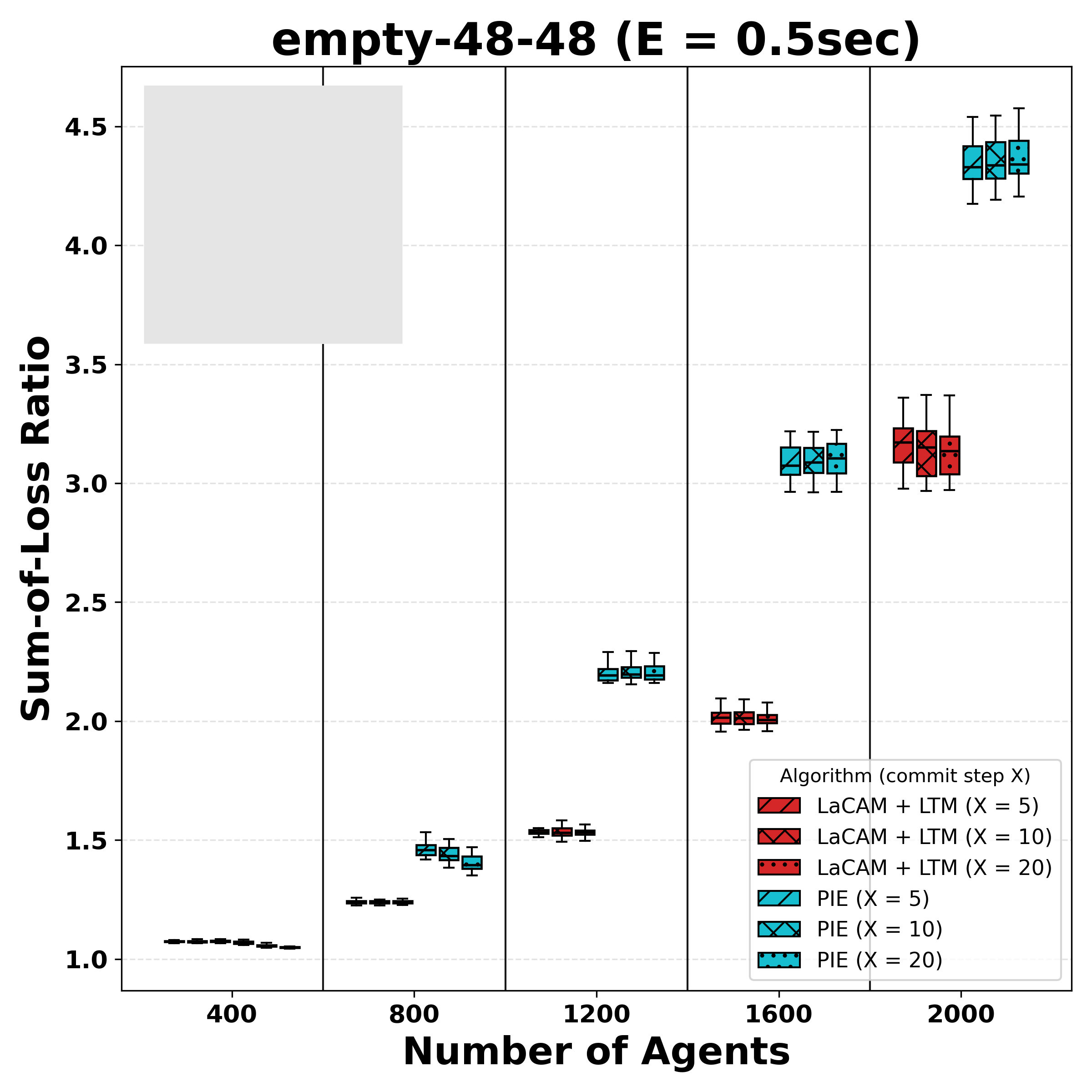}
\end{minipage}&

\hfill

\begin{minipage}{.24\linewidth}
\centering
\includegraphics[scale=0.21]{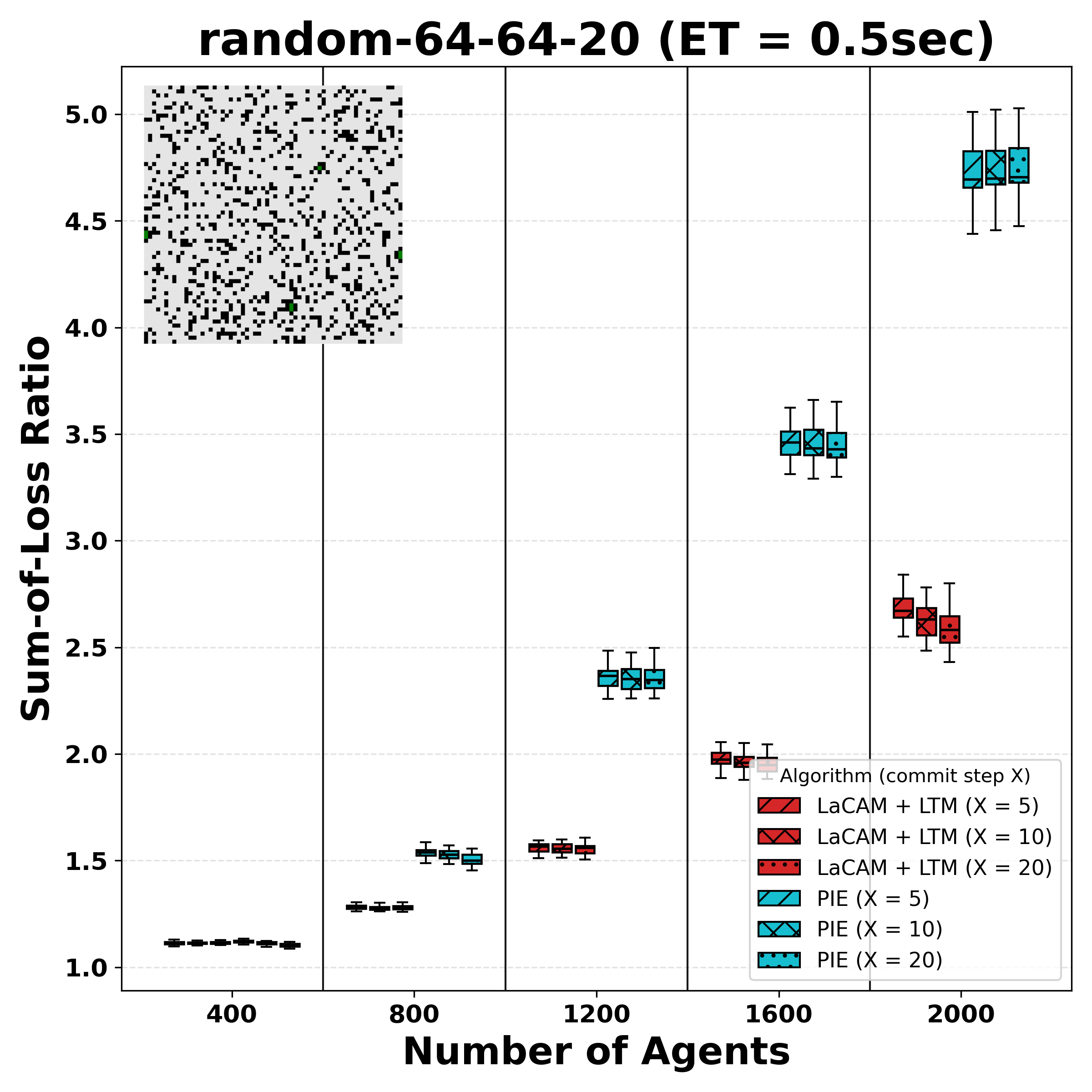}
\end{minipage}
\hfill
\begin{minipage}{.24\linewidth}
\centering
\includegraphics[scale=0.21]{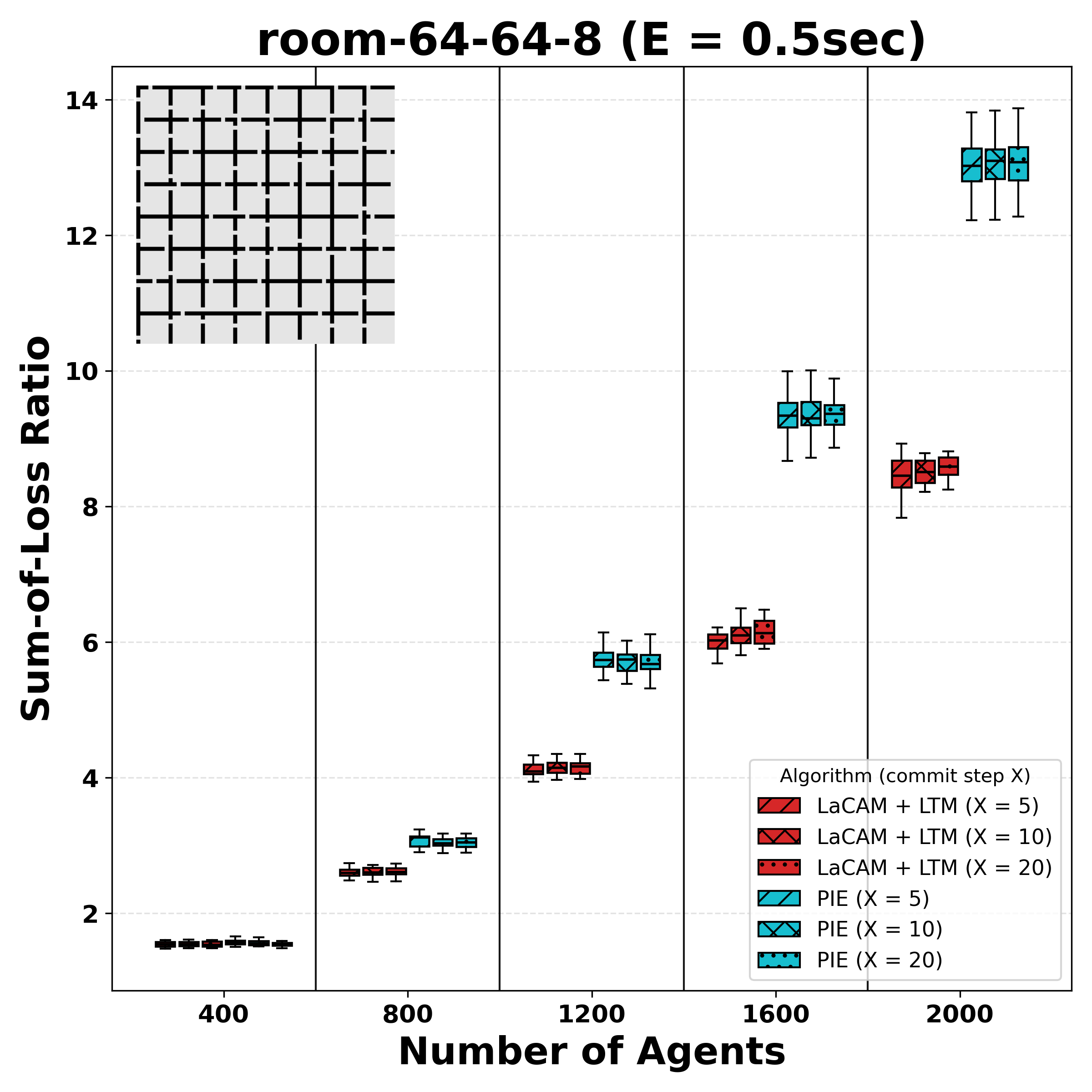}
\end{minipage}

\hfill
\begin{minipage}{.24\linewidth}
\centering
\includegraphics[scale=0.21]{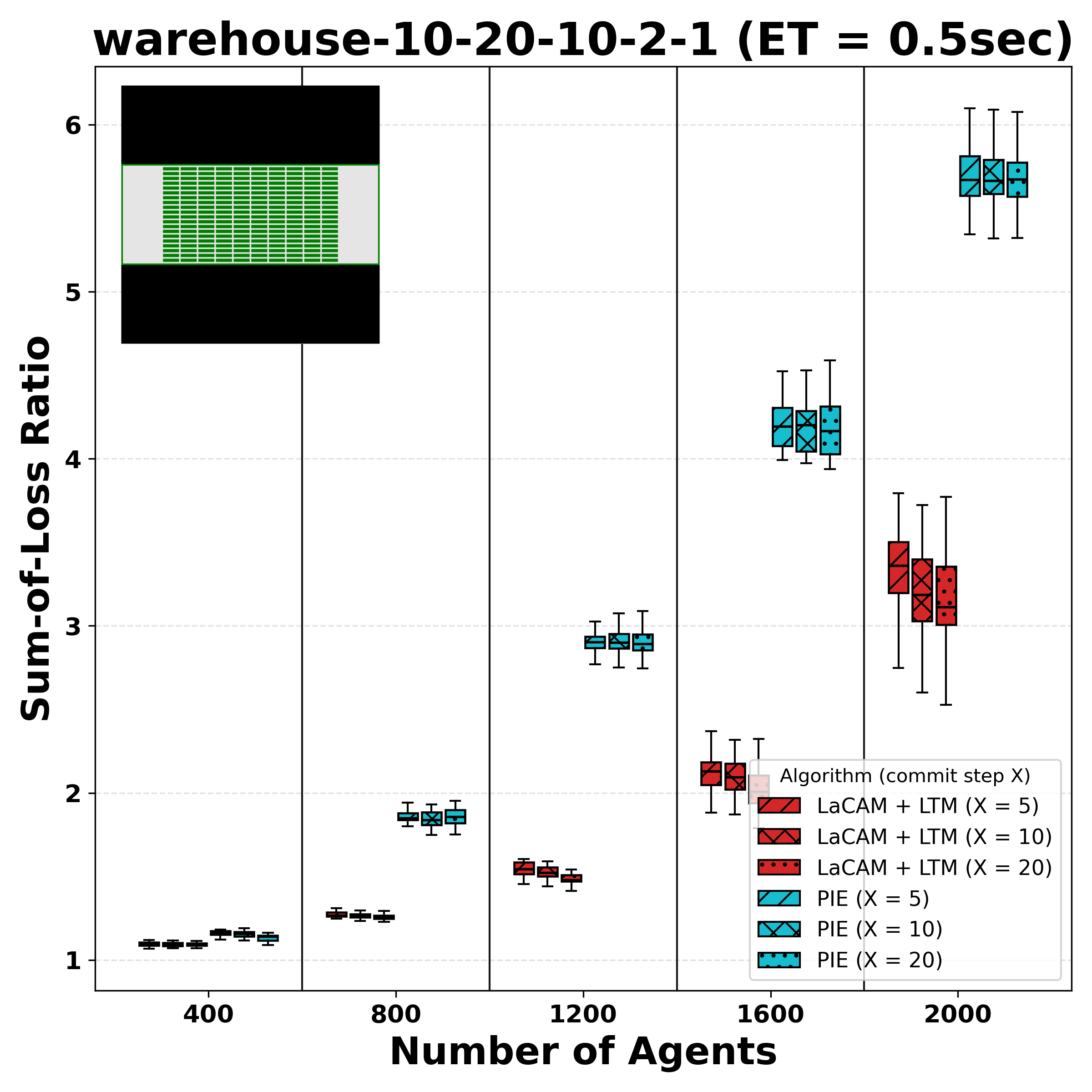}
\end{minipage}
\end{tabular}

\caption{ Planning-and-execution MAPF, each figure reports the sum of loss ratio for LaCAM*, LaCAM*+TO, LaCAM*+SUO, and the proposed LaCAM+LTM. Results are shown for different execution times (E = 0.1 s and 0.5 s) and commitment steps of 5, 10, and 20 steps, respectively. Values are averaged over instances; error bars indicate standard deviation. Lower values indicate better solution quality.}
\label{fig::execution}
\end{figure*}

\subsection{One-shot MAPF}
We evaluate the one-shot MAPF setting on eight grid-based maps from the standard MAPF benchmark sourced from~\cite{stern2019multi}. To assess scalability, we generate 25 random instances per map and vary the number of agents from small cases to 2000 agents. All algorithms have a runtime limit of 30 seconds.

We compare LaCAM+LTM against three representative baselines.
(1) \textbf{LaCAM*}~\cite{okumura2023lacam}, the anytime variant of LaCAM.
(2) \textbf{LaCAM*+TO}, which integrates traffic optimisation~\cite{chen2024traffic} by repeatedly running Frank–Wolfe optimisation to compute guidance paths that minimise a handcrafted traffic objective. These guidance paths are then used to guide PIBT in LaCAM*. 
(3) \textbf{LaCAM*+SUO}, which applies space utilization optimization~\cite{okumura2023engineering} to minimize conflicts among guidance paths. 
For LaCAM*, TO and SUO, we use the authors' original implementation~\cite{okumura2023improving,okumura2023engineering,chen2024traffic}.
For our implementation, \textbf{LaCAM*+LTM} is built on top of the original LaCAM* codebase. To ensure fair comparison, traffic optimisation is run for 15 seconds to allow the Frank–Wolfe procedure to converge before executing LaCAM*. SUO is run until the guidance paths stabilise, following the authors’ recommendations.

\subsubsection{Solution Quality} 
Figure~\ref{fig::solution_quality} compares solution quality under a 30-second time limit using the sum-of-loss ratio, computed relative to a trivial lower bound given by the sum of individual shortest-path costs. Overall, the proposed LaCAM*+LTM consistently achieves lower loss ratios than the baselines, particularly in dense scenarios with many agents, indicating better scalability as congestion increases.
While LaCAM* performs competitively for small agent counts, its solution quality degrades rapidly as density grows due to its depth-first search strategy, which limits improvement as makespan increases. LaCAM*+TO and LaCAM*+SUO partially mitigate this issue by computing guidance paths at the start of the search; however, since these paths remain fixed, their effectiveness diminishes in large-scale instances.
In contrast, LaCAM*+LTM combines frequent restarts with an adaptive Lightweight Traffic Map that continuously captures congestion patterns from previous iterations and updates heuristic guidance. This self-adaptive mechanism enables LaCAM*+LTM to efficiently generate higher-quality solutions over time without handcrafted optimization objectives. 

Figure~\ref{fig::coverage_plot_64x64_1000} further demonstrates its superior anytime performance, showing faster convergence and a larger number of returned solutions compared to competing methods.
We do not directly compare against hybrid approaches such as LaCAM3~\cite{okumura2023lacam}, as their performance largely depends on parallelization and LNS integration. When parallelism is disabled, LaCAM3 becomes equivalent  to LaCAM*+SUO. In the next section, we further justify the advantages of our approach through a comparison with an equivalent single-thread solver (PIE) in the context of planning and execution.

\subsubsection*{Inside the Lightweight Traffic Map} Figure~\ref{fig::traffic_map} illustrates how the Lightweight Traffic Map (LTM) evolves during the search. In the early stages of LaCAM*+LTM, the LTM captures congestion induced by the first few solutions, with traffic initially spread across the map and only a small number of emerging high-traffic corridors. The resulting increases in traversal costs are then used to guide the low-level configuration generator, PIBT, producing improved heuristics in subsequent iterations and enabling LaCAM*+LTM to iteratively enhance solution quality while still returning solutions immediately. As the search progresses, congestion information from successive iterations accumulates: frequently congested area become increasingly penalized, leading to clearer and more concentrated congestion patterns along narrow passages and shared routes. This adaptive redistribution encourages agents to diversify their paths, alleviates congestion in heavily used regions, and ultimately results in improved solution quality over time.








\subsection{Planning and Execution MAPF}
We conduct experiments on the same benchmark maps used in the one-shot MAPF setting and compare against the state-of-the-art algorithm PIE~\cite{zhang2024planning}. PIE can be viewed as a hybrid approach that combines LaCAM*~\cite{okumura2023lacam} for quickly generating an initial feasible solution and LNS~\cite{mapf-lns2} for improving partial solutions. Specifically, PIE first uses LaCAM* to obtain a feasible plan for fast action commitment. During each subsequent planning window, it then applies LNS to refine the partial plan from the last committed configuration toward the goal, repeating this process until all agents reach their goals.
To evaluate performance under different execution constraints, we vary the execution time $E$ among 0.1 s, and 0.5 s, and the commitment horizon $X$ among 5, 10, and 20 steps.

Figure~\ref{fig::execution} compares solution quality between the proposed LaCAM*+LTM and PIE under different execution constraints. Overall, LaCAM*+LTM consistently achieves lower sum-of-loss ratios as the number of agents increases, demonstrating stronger robustness in dense scenarios. When execution time is short (ET = 0.1 s), both methods are constrained by limited planning time; nevertheless, LaCAM*+LTM maintains better solution quality. As execution time increases to 0.5s, the performance gap widens, indicating that LaCAM*+LTM can more effectively exploit additional planning time.
In contrast, PIE degrades more rapidly as agent density increases, mainly due to limitations of LNS. First, LNS improves solutions through destroy-and-repair operations that rely on space-time A*, which becomes increasingly expensive in high-density scenarios, reducing the number of improvement iterations. Moreover, LNS typically replans only a small subset of agents, which limits the impact of each iteration in large-scale instances. Second, in the planning-and-execution setting, LNS initially replans long paths from the start to the goal, making each iteration costly; as execution progresses, although replanned paths become shorter, the remaining room for improvement also diminishes. Consequently, local refinement alone is insufficient to resolve global congestion patterns.
By contrast, LaCAM+LTM continuously updates the traffic map and restarts from the current configuration, enabling adaptive global guidance that mitigates persistent congestion and scales better than PIE.
\section{Conclusion}
We proposed the Lightweight Traffic Map (LTM), an online and low-overhead mechanism that improves the anytime performance of LaCAM* by dynamically capturing congestion during search. By continuously updating traffic-aware guidance without offline optimization or handcrafted objectives, LaCAM*+LTM achieves faster convergence and higher solution quality than existing guidance-based methods. Experiments in both one-shot and planning-and-execution settings demonstrate its superior scalability and robustness in dense MAPF scenarios.

%% file: ijcai26.bib
@inproceedings{stern2019multi,
  title={Multi-agent pathfinding: Definitions, variants, and benchmarks},
  author={Stern, Roni and Sturtevant, Nathan and Felner, Ariel and Koenig, Sven and Ma, Hang and Walker, Thayne and Li, Jiaoyang and Atzmon, Dor and Cohen, Liron and Kumar, TK and others},
  booktitle={Proceedings of the International Symposium on Combinatorial Search},
  volume={10},
  number={1},
  pages={151--158},
  year={2019}
}

@inproceedings{okumura2023lacam,
  title={Lacam: Search-based algorithm for quick multi-agent pathfinding},
  author={Okumura, Keisuke},
  booktitle={Proceedings of the AAAI Conference on Artificial Intelligence},
  volume={37},
  number={10},
  pages={11655--11662},
  year={2023}
}

@inproceedings{okumura2023improving,
  title={Improving LaCAM for scalable eventually optimal multi-agent pathfinding},
  author={Okumura, Keisuke},
  booktitle={Proceedings of the Thirty-Second International Joint Conference on Artificial Intelligence},
  pages={243--251},
  year={2023}
}

@inproceedings{okumura2023engineering,
  title = {Engineering LaCAM$^\ast$: Towards Real-Time, Large-Scale, and Near-Optimal Multi-Agent Pathfinding},
  booktitle={Proceedings of International Conference on Autonomous Agents and Multiagent Systems (AAMAS)},
  year={2024},
  author={Okumura, Keisuke}
}

@inproceedings{chen2024traffic,
  title={Traffic flow optimisation for lifelong multi-agent path finding},
  author={Chen, Zhe and Harabor, Daniel and Li, Jiaoyang and Stuckey, Peter J},
  booktitle={Proceedings of the AAAI Conference on Artificial Intelligence},
  volume={38},
  number={18},
  pages={20674--20682},
  year={2024}
}

@article{wurman2008coordinating,
  author     = {Peter R. Wurman and
                Raffaello D'Andrea and
                Mick Mountz},
  bibsource  = {dblp computer science bibliography, https://dblp.org},
  doi        = {10.1609/AIMAG.V29I1.2082},
  journal    = {{AI} Mag.},
  number     = {1},
  pages      = {9--20},
  title      = {{C}oordinating {H}undreds of {C}ooperative, {A}utonomous {V}ehicles in {W}arehouses},
  url        = {https://doi.org/10.1609/aimag.v29i1.2082},
  volume     = {29},
  year       = {2008}
}

@inproceedings{silver2005cooperative,
  author     = {David Silver},
  bibsource  = {dblp computer science bibliography, https://dblp.org},
  booktitle  = {Proceedings of the First Artificial Intelligence and Interactive Digital
                Entertainment Conference, June 1-5, 2005, Marina del Rey, California,
                {USA}},
  editor     = {R. Michael Young and
                John E. Laird},
  pages      = {117--122},
  publisher  = {{AAAI} Press},
  title      = {{C}ooperative {P}athfinding},
  year       = {2005}
}

@inproceedings{flatland,
  author     = {Jiaoyang Li and
                Zhe Chen and
                Yi Zheng and
                Shao{-}Hung Chan and
                Daniel Harabor and
                Peter J. Stuckey and
                Hang Ma and
                Sven Koenig},
  bibsource  = {dblp computer science bibliography, https://dblp.org},
  booktitle  = {Proceedings of the Thirty-First International Conference on Automated
                Planning and Scheduling},
  editor     = {Susanne Biundo and
                Minh Do and
                Robert Goldman and
                Michael Katz and
                Qiang Yang and
                Hankz Hankui Zhuo},
  pages      = {477--485},
  publisher  = {{AAAI} Press},
  title      = {{S}calable {R}ail {P}lanning and {R}eplanning: {W}inning the 2020 {F}latland {C}hallenge},
  year       = {2021}
}

@article{li2021eecbs,
  author       = {Jiaoyang Li and
                  Wheeler Ruml and
                  Sven Koenig},
  title        = {{EECBS:} {A} Bounded-Suboptimal Search for Multi-Agent Path Finding},
  booktitle    = {Thirty-Fifth {AAAI} Conference on Artificial Intelligence},
  pages        = {12353--12362},
  publisher    = {{AAAI} Press},
  year         = {2021},
  url          = {https://doi.org/10.1609/aaai.v35i14.17466},
  doi          = {10.1609/AAAI.V35I14.17466},
  bibsource    = {dblp computer science bibliography, https://dblp.org}
}

@inproceedings{mapf-lns2,
  author     = {Jiaoyang Li and
                Zhe Chen and
                Daniel Harabor and
                Peter J. Stuckey and
                Sven Koenig},
  bibsource  = {dblp computer science bibliography, https://dblp.org},
  booktitle  = {Thirty-Sixth {AAAI} Conference on Artificial Intelligence},
  pages      = {10256--10265},
  publisher  = {{AAAI} Press},
  title      = {{MAPF-LNS2:} {F}ast {R}epairing for {M}ulti-Agent {P}ath {F}inding via {L}arge {N}eighborhood {S}earch},
  year       = {2022}
}

@inproceedings{cluster_reasoning,
  author     = {Bojie Shen and
                Zhe Che and
                Jiaoyang Li and
                Muhammad Aamir Cheema and
                Daniel Damir Harabor and
                Peter J. Stuckey},
  bibsource  = {dblp computer science bibliography, https://dblp.org},
  booktitle  = {Proceedings of the Thirty-Third International Conference on Automated
                Planning and Scheduling, July 8-13, 2023, Prague, Czech Republic},
  doi        = {10.1609/ICAPS.V33I1.27217},
  editor     = {Sven Koenig and
                Roni Stern and
                Mauro Vallati},
  pages      = {384--392},
  publisher  = {{AAAI} Press},
  title      = {{B}eyond {P}airwise {R}easoning in {M}ulti-Agent {P}ath {F}inding},
  url        = {https://doi.org/10.1609/icaps.v33i1.27217},
  year       = {2023}
}

@article{okumura2022priority,
  author       = {Keisuke Okumura and
                  Manao Machida and
                  Xavier D{\'{e}}fago and
                  Yasumasa Tamura},
  title        = {Priority inheritance with backtracking for iterative multi-agent path
                  finding},
  journal      = {Artif. Intell.},
  volume       = {310},
  pages        = {103752},
  year         = {2022},
  url          = {https://doi.org/10.1016/j.artint.2022.103752},
  doi          = {10.1016/J.ARTINT.2022.103752},
  bibsource    = {dblp computer science bibliography, https://dblp.org}
}

@inproceedings{JiangSoCS24,
  author    = "He Jiang and Yulun Zhang and Rishi Veerapaneni and Jiaoyang Li",
  title     = "Scaling Lifelong Multi-Agent Path Finding to More Realistic Settings: Research Challenges and Opportunities",
  booktitle = "Proceedings of the Symposium on Combinatorial Search",
  pages     = "234-242",
  year      = "2024",
  doi       = "10.1609/socs.v17i1.31565",
}

@article{SUO,
  author       = {Shuai D. Han and
                  Jingjin Yu},
  title        = {Optimizing Space Utilization for More Effective Multi-Robot Path Planning},
  booktitle    = {2022 International Conference on Robotics and Automation, {ICRA} 2022},
  pages        = {10709--10715},
  publisher    = {{IEEE}},
  year         = {2022},
}

@inproceedings{OnlineGGO,
  author       = {Hongzhi Zang and
                  Yulun Zhang and
                  He Jiang and
                  Zhe Chen and
                  Daniel Harabor and
                  Peter J. Stuckey and
                  Jiaoyang Li},
  editor       = {Toby Walsh and
                  Julie Shah and
                  Zico Kolter},
  title        = {Online Guidance Graph Optimization for Lifelong Multi-Agent Path Finding},
  booktitle    = {Thirty-Ninth {AAAI} Conference on Artificial Intelligence},
  pages        = {14726--14735},
  publisher    = {{AAAI} Press},
  year         = {2025},
}

@inproceedings{GGO,
  author    = "Yulun Zhang and He Jiang and Varun Bhatt and Stefanos Nikolaidis and Jiaoyang Li",
  title     = "Guidance Graph Optimization for Lifelong Multi-Agent Path Finding",
  booktitle = "Proceedings of the International Joint Conference on Artificial Intelligence",
  pages     = "311-320",
  year      = "2024",
  doi       = "10.24963/ijcai.2024/35",
}

@article{shen2023tracking,
  author     = {Bojie Shen and
                Zhe Chen and
                Muhammad Aamir Cheema and
                Daniel Damir Harabor and
                Peter J. Stuckey},
  bibsource  = {dblp computer science bibliography, https://dblp.org},
  doi        = {10.48550/ARXIV.2305.08446},
  eprint     = {2305.08446},
  eprinttype = {arXiv},
  journal    = {CoRR},
  title      = {{T}racking {P}rogress in {M}ulti-Agent {P}ath {F}inding},
  url        = {https://doi.org/10.48550/arXiv.2305.08446},
  volume     = {abs/2305.08446},
  year       = {2023}
}

@article{bcp-mapf,
  author       = {Edward Lam and
                  Pierre Le Bodic and
                  Daniel Harabor and
                  Peter J. Stuckey},
  title        = {Branch-and-cut-and-price for multi-agent path finding},
  journal      = {Comput. Oper. Res.},
  volume       = {144},
  pages        = {105809},
  year         = {2022},
  url          = {https://doi.org/10.1016/j.cor.2022.105809},
  doi          = {10.1016/J.COR.2022.105809},
  bibsource    = {dblp computer science bibliography, https://dblp.org}
}

@inproceedings{lazycbs,
  author       = {Graeme Gange and
                  Daniel Harabor and
                  Peter J. Stuckey},
  editor       = {J. Benton and
                  Nir Lipovetzky and
                  Eva Onaindia and
                  David E. Smith and
                  Siddharth Srivastava},
  title        = {Lazy {CBS:} Implicit Conflict-Based Search Using Lazy Clause Generation},
  booktitle    = {Proceedings of the Twenty-Ninth International Conference on Automated
                  Planning and Scheduling},
  pages        = {155--162},
  publisher    = {{AAAI} Press},
  year         = {2019},
  url          = {https://ojs.aaai.org/index.php/ICAPS/article/view/3471},
  bibsource    = {dblp computer science bibliography, https://dblp.org}
}

@inproceedings{zhang2024planning,
  title={Planning and execution in multi-agent path finding: Models and algorithms},
  author={Zhang, Yue and Chen, Zhe and Harabor, Daniel and Le Bodic, Pierre and Stuckey, Peter J},
  booktitle={Proceedings of the International Conference on Automated Planning and Scheduling},
  volume={34},
  pages={707--715},
  year={2024}
}
